\documentclass[acmlarge, manuscript, nonacm]{acmart}

\usepackage[utf8]{inputenc}
\usepackage{ulem}
\usepackage{tikz}
\usetikzlibrary{arrows.meta, positioning}

\usepackage{threeparttable}
\usepackage{multirow}

\usepackage{booktabs}
\usepackage{pifont}
\usepackage{xcolor}
\usepackage{graphicx}
\usepackage{xspace}
\usepackage{tabularx}
\usepackage{listings}
\usepackage{longtable}

\usepackage[table]{xcolor}

\newcommand{\cmark}{\textcolor{green!70!black}{\ding{51}}} 
\newcommand{\xmark}{\textcolor{red!70!black}{\ding{55}}}   

\AtBeginDocument{%
  }

\begin{document}



\newcommand\bnote[1]{\textcolor{blue}{#1}}
\newcommand\note[1]{\textcolor{red}{#1}}

\newcommand{\sys}{\textsc{\textbf{ProcAgent}}\xspace}

\title{\sys~: An Agentic Framework for Procedural Task Guidance on Edge with Human-in-the-Loop}



\author{Azizul Zahid}
\affiliation{%
  \institution{University of Tennessee Knoxville}
  \city{Knoxville}
  \country{USA}}
\email{azahid@vols.utk.edu}

\author{Subrata Biswas}
\affiliation{%
  \institution{Worcester Polytechnic Institute}
  \city{Worcester}
  \country{USA}}
\email{sbiswas@wpi.edu}

\author{Bashima Islam}
\authornote{Equal supervision.}
\affiliation{%
  \institution{Worcester Polytechnic Institute}
  \city{Worcester}
  \country{USA}}
\email{bislam@wpi.edu}

\author{Sai Swaminathan}
\authornotemark[1]
\email{sswamin6@utk.edu}
\affiliation{%
  \institution{University of Tennessee Knoxville}
  \city{Knoxville}
  \country{USA}}






\renewcommand{\shortauthors}{Zahid et al.}

\begin{abstract}

Procedural tasks such as furniture assembly and home repair impose substantial cognitive demands because users must interpret instructions, track task progress, reason about spatial state, and recover from errors while performing physical actions. Prior multimodal assistants have shown promise for procedural guidance, but most rely on cloud inference and fixed always-on perception, making them poorly suited to privacy-sensitive, latency-critical domestic settings. We present \textbf{\sys}, a fully on-device, agentic, vision-based procedural assistant for real-time adaptive guidances on a single NVIDIA Jetson AGX Orin. \sys~uses a propose-and-verify architecture that combines low-latency continuous perception, a symbolic task graph, on-demand vision-language verification, and an LLM-based interaction agent. The system continuously proposes user progress, invokes expensive visual reasoning only when ambiguity or likely deviation arises, and supports both reactive question answering and proactive intervention with human-in-the-loop confirmation. We evaluate \sys~along four dimensions: perception accuracy, reasoning, task-level performance, and user experience. Despite running entirely on-device, the system maintains responsive interaction, resolving text-only queries in approximately 2 seconds and visually grounded queries in approximately 8 seconds. In a user study with 10 participants completing assembly tasks, \sys~receives positive ratings for comprehensibility, actionability, and privacy comfort. These results show that adaptive procedural assistance can be achieved entirely on edge hardware without sacrificing usability.

\end{abstract}





\maketitle



\section{Introduction}

\begin{figure}[!htb]
    \centering
    \includegraphics[trim=0 100 0 60, clip, width=\textwidth]{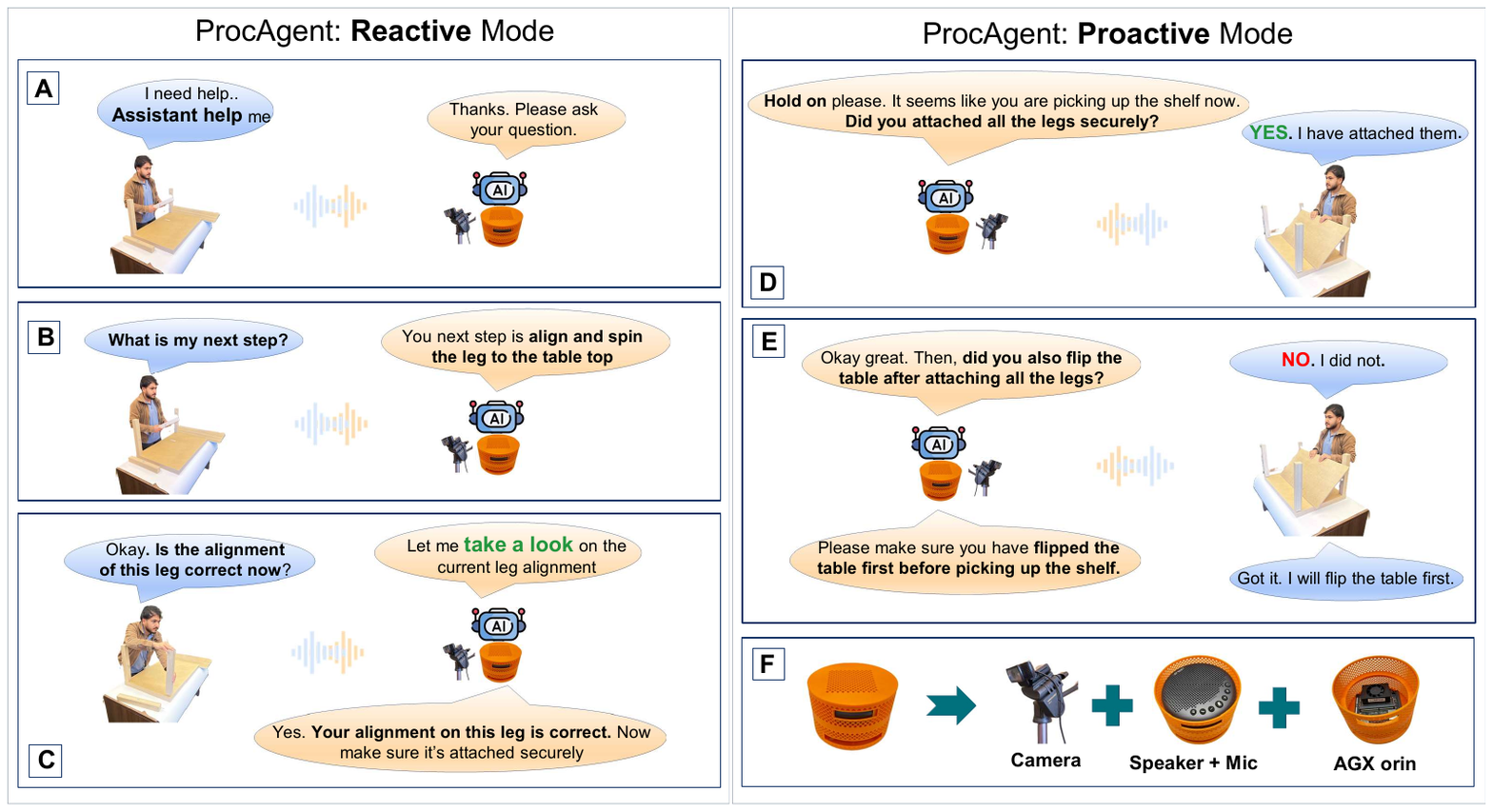}
    \caption{\sys, deployed on home assitant device, supports furniture assembly in both reactive and proactive modes. In the \textbf{reactive mode (A-C)}, the user explicitly requests help, asks for the next assembly step, and receives visual verification of leg alignment. In the \textbf{proactive mode} (D-E), \sys monitors the task and intervenes when error occurs to confirm whether the legs are securely attached and whether the table has been flipped before lifting, thereby preventing procedural mistakes. (F) shows the \sys's deployment on home assistant device with it's hardware, including a camera, speaker and microphone, and an NVIDIA Jetson AGX Orin module for perception, interaction, and computation.}
    \label{fig:teasure}
    \Description[]{}
    \vspace{-2mm}
\end{figure}

Procedural tasks, such as furniture assembly, home repair, laboratory protocols, craft fabrication, share a simple structure: a sequence of discrete physical actions that transform components into a finished artifact. Yet executing them imposes substantial cognitive demand. The user must simultaneously parse instructions, maintain a mental model of current state, locate parts, plan the next action, and monitor for errors, all while performing the physical manipulation itself~\cite{eesee2025impact}. This concurrent load on working memory, spatial reasoning, and attention is especially taxing for non-experts~\cite{sewell2016measuring}. This demand is not uniform across the task. A user executing a straightforward step needs no assistance; a user stuck on an ambiguous diagram at step 17 of 38 needs immediate, spatially grounded guidance. Static manuals cannot make this distinction, and AR overlays or pre-recorded videos, though richer, remain non-adaptive: they cannot detect errors, recover from deviations, or answer free-form questions grounded in the current workspace state~\cite{hartanto2019development}. What is needed is a just-in-time assistant that monitors progress, detects confusion, and intervenes only when warranted.

A growing body of work has begun to address this need. Recent datasets such as ProMQA-Assembly~\cite{hasegawa2025promqa} and CaptainCook4D~\cite{peddi2024captaincook4d} have established procedural QA and error-detection benchmarks, while systems like PrISM-Q\&A~\cite{riku2024imwut} demonstrate step-aware voice assistants that pair multimodal procedure tracking with LLMs, and CHEF-VL~\cite{wang2025imwut} shows that vision-language models can detect cognitive sequencing errors in cooking. Other efforts ground task assistance in single demonstrations or push real-time action recognition onto embedded platforms~\cite{meng2008real, luo2025edgeoar}. Collectively, these results validate that LLM- and VLM-driven procedural assistants are both feasible and useful.
Yet two limitations cut across this prior work. First, most systems rely on cloud inference for the LLM or VLM components~\cite{meng2008real}, which is problematic for a vision-based assistant that must continuously observe the user's workspace. Streaming this video to remote servers raises well-documented privacy and ``surveillance anxiety'' concerns in domestic settings~\cite{dourish2004we}, introduces 5--10 second round-trip latencies that break the just-in-time coaching loop, and makes the system unusable in network-constrained environments. Second, procedural tasks are inherently multimodal: some turns require only language reasoning, others visual grounding, others knowledge retrieval, and many require all three. Existing systems treat perception as either always-on, incurring continuous compute and thermal cost, or as a fixed pipeline applied uniformly to every turn regardless of complexity. Neither scales to sustained real-time interaction.

What is missing is an \emph{agentic} architecture that dynamically selects the appropriate tool per turn -- invoking vision only when spatial verification is needed, retrieving knowledge only when procedural detail is required, and otherwise relying on language reasoning alone, and that does so entirely on-device. Achieving this on commodity edge hardware (e.g., NVIDIA Jetson-class devices) is non-trivial: continuous billion-parameter VLM inference exhausts unified memory and triggers thermal throttling within minutes, while lightweight detectors lack the semantic granularity to reliably distinguish fine-grained procedural states such as a partially aligned leg from a fully secured one~\cite{huang2025litevlm}. The architectural question, then, is not whether LLM agents \emph{can} assist with procedural tasks, prior work has shown they can, but how to deliver agentic intelligence under the joint constraints of privacy, latency, and edge compute.

We address these constraints with \textbf{\sys}, a fully on-device, agentic, vision-based procedural assistant for real-time adaptive coaching in domestic environments. \sys~runs entirely on a single NVIDIA Jetson AGX Orin and is built around a \emph{propose-and-verify} paradigm. A lightweight perception module, initialized through contrastive image pretraining and fine-tuned on a small set of task-specific step descriptions, runs continuously and proposes what it believes the user is doing. These proposals are treated as tentative hypotheses and do not directly update the system’s understanding of progress. Instead, a symbolic task representation, an explicit graph of steps and their ordering rules, determines which proposals are plausible and which require confirmation. When needed, a vision--language model is invoked on demand to verify the current state, allowing the system to use high-fidelity visual reasoning only at critical moments rather than continuously. An LLM-based agent handles interaction with the user, deciding when to intervene, what to ask, and when to defer to the user’s judgment. As illustrated in Figure~1, \sys~supports both \emph{reactive} and \emph{proactive} interaction modes. In reactive mode, the agent responds to user-initiated queries, providing step guidance or verifying actions on demand. In proactive mode, the agent continuously monitors task progress and intervenes only when it detects a likely deviation from the expected sequence, using brief confirmation to correct errors before they propagate. Together, these components decouple \emph{perceptual tracking} from \emph{semantic reasoning}, allowing each to operate at the latency and fidelity appropriate to its role. The agent further reasons about when visual grounding is necessary before invoking vision models (e.g., answering “what is my next step?” from the task model while reserving vision for “is this aligned correctly?”). In addition, \textbf{\sys} incorporates a human-in-the-loop confirmation strategy that defers to the user when perception and procedural structure disagree (e.g., briefly asking for confirmation before correcting a suspected mistake). This design reflects a key interaction principle: in voluntary-use settings, the cost of a false intervention often exceeds that of a missed one.

We evaluate \sys~along four dimensions: perception accuracy, reasoning, task-level performance, and user experience. On a single NVIDIA Jetson AGX Orin, the system operates entirely on-device and supports real-time interaction, with responses resolving in approximately 2 seconds for text-only queries and 8 seconds for visually grounded reasoning. Across agent backbones, we observe clear trade-offs between response quality and tool-use reliability, with our selected model providing the most consistent performance. In a user study with 10 participants performing assembly tasks, \sys~receives broadly positive ratings for comprehensibility, actionability, and privacy comfort, with participants explicitly valuing the system’s on-device operation. Together, these results demonstrate that adaptive, agentic procedural assistance can be realized entirely on edge hardware without sacrificing usability, suggesting that the trade-off between privacy, latency, and capability is architectural rather than fundamental. Our work makes the following contributions:

\textbf{1. An end-to-end, fully on-device agentic system for procedural assistance.}

We present \textbf{\sys}, a vision-based procedural assistant that integrates perception, structured task tracking, and language-based interaction into a unified system that operates entirely on edge hardware. We demonstrate that real-time, adaptive procedural guidance can be achieved without cloud inference while preserving usability and user trust.

\textbf{2. A framework for making multimodal procedural assistants efficient and reliable on edge devices.} 
Instead of running perception and reasoning continuously, we show how lightweight, always-on model can continuously propose the user’s actions and invoke expensive visual reasoning only when needed, while task structure and user confirmation maintain correctness under our proposed framework.

\textbf{3. An empirical characterization of performance and user interaction in on-device procedural assistants.}

Through system ablations and a user study, we show how different components contribute to perception accuracy, procedural correctness, resource efficiency, and user experience. Our results identify human-in-the-loop confirmation as a key driver of sequence adherence and demonstrate that on-device deployment meaningfully improves user perceptions of privacy and trust.

\section{Related Work}



\subsection{Vision-Based Assistants}


Early vision-based assistants relied on fiducial markers, depth sensors, or structured lighting to track part positions and infer assembly state in controlled manufacturing environments\cite{re2016impact,radkowski2015augmented}. While these systems demonstrated the feasibility of vision-driven guidance, their dependence on instrumented environments and specialized hardware limited their applicability to real-world, unstructured settings.

The emergence of deep learning brought significant advances in visual recognition for procedural tasks. Convolutional neural network-based approaches enabled action recognition and step detection from raw video without environment instrumentation, with datasets such as the IKEA Assembly Dataset\cite{ben2021ikea} and Assembly101\cite{sener2022assembly101} providing large-scale benchmarks for evaluating assembly-specific recognition. Egocentric video understanding, explored through datasets like EPIC-Kitchens\cite{damen2018scaling} and WEAR\cite{bock2024wear} further demonstrated that first-person visual perspectives could capture the fine-grained hand-object interactions central to procedural task performance. These advances established that visual recognition models could reliably identify what a user was doing, but not what they should do next, nor how to communicate that to them.

More recent systems have begun to bridge perception and guidance by pairing visual recognition with natural language generation. AR-based assistants such as HoloAssist\cite{wang2023holoassist} combine egocentric video understanding with step-level instruction delivery, while systems built on large vision-language models like InstructBLIP\cite{dai2023instructblip} and LLaVA\cite{liu2023visual} have shown impressive ability to answer open-ended visual queries about procedural scenes. In the surgical and medical domain, vision-based guidance systems have demonstrated that real-time visual feedback can meaningfully reduce procedural errors\cite{gao2021trans, kletz2019identifying}. Despite these advances, a consistent limitation persists across the literature: visual recognition and language generation are treated as loosely coupled modules rather than components of a unified, state-aware system. The vision module observes, and the language module responds, but neither maintains a persistent understanding of where the user is in the task, what has already been verified, or what constraints govern what can legally happen next.

\subsection{Error Detection and Recovery Systems in Procedural Task}

Detecting and recovering from errors in procedural tasks has been a longstanding challenge across robotics, human-computer interaction, and intelligent tutoring systems. Early approaches relied on rule-based monitors that compared sensor readings or symbolic state representations against predefined correct sequences, flagging deviations when observed behavior diverged from the expected plan\cite{hegemann2022learning}. While effective in highly controlled environments, these systems were brittle in the face of real-world variability. Partial occlusions, user hesitation, and ambiguous intermediate states were enough to cause cascading misclassifications.

The introduction of video-based action recognition brought greater flexibility to error detection in procedural settings. Systems such as Assembly101\cite{sener2022assembly101} and FineGym\cite{shao2020finegym} demonstrated that fine-grained temporal action segmentation could reliably identify when a step was performed incorrectly or out of order. Similarly, error detection in cooking and surgical domains\cite{lea2017temporal, gao2021trans} showed that temporal convolutional networks and transformer-based architectures could capture the sequential dependencies between steps well enough to flag anomalies without explicit rule encoding. CHEF-VL\cite{wang2025imwut} proposes a framework for cooking scenarios that detects sequence errors by leveraging two vision-language models (VLMs): one for human activity recognition and the other for tracking state transitions. Additionally, it introduces an action-state merging function to reduce prediction noise. However, these systems operate purely as passive detectors, they identify that something has gone wrong but provide no mechanism for communicating the error to the user or guiding them toward recovery.

Additionally, LLM-based chatbots\cite{tang2025imwut} are conversational, either they retrieved context from knowledge base or the user interactions. They can not be applied in multimodal agentic settings for procedural tasks where agent needs to understand the sequential progress of the user and give adaptive feedback to recover from error if user deviates. CataractBot\cite{ramjee2025imwut} is LLM-based chatbot that answers patient questions, provide expert-in-the-loop medical guidance, and support care-related dialogue, but they do not maintain a step-aware representation of an ongoing procedure. As a result, they cannot detect whether a user has followed the correct sequence, identify illegal transitions in real time, or trigger recovery based on the current process state.

More recent work has begun to close this gap by coupling error detection with natural language feedback. InstructBLIP\cite{dai2023instructblip} and similar vision-language systems have been used to generate post-hoc error explanations from visual observations, while systems like HELPER\cite{sarch2023open} combine LLM-based planning with error recovery in embodied navigation tasks. In the procedural assembly domain, specifically, work by Ben Shabat et al.\cite{ben2021ikea} on the IKEA Assembly Dataset introduced benchmarks for step recognition and order verification, laying important groundwork for systems that reason about assembly correctness. 

Nevertheless, these approaches share a common limitation: error recovery is treated as a separate, reactive process that is triggered only after a mistake has already been made and confirmed, rather than being integrated into a continuous, proactive guidance loop.
The gap our work addresses is the absence of a unified system that combines real-time error detection, procedural constraint enforcement, and natural language recovery within a single closed-loop agent. Existing systems either detect errors without communicating them, communicate without grounding responses in verified visual state, or recover without preventing the same error from recurring. 


\subsection{LLM-Based Agentic Systems for Procedural Task Guidance}

LLM-based agents have expanded procedural task guidance by combining language reasoning with external tools and perception. Early frameworks such as ReAct\cite{yao2023reactsynergizingreasoningacting} and Toolformer\cite{schick2023toolformerlanguagemodelsteach} showed that language models can coordinate tools to solve multi-step tasks, while later embodied systems extended these ideas to settings that require state tracking, action planning, and real-time adaptation\cite{ahn2022icanisay, hu2023look}.

For procedural assistance, prior systems have combined language and vision to support tasks such as cooking, maintenance, and manufacturing\cite{zhao2025guided}. Systems such as PrISM-Q\&A\cite{riku2024imwut} and 3D-GPT\cite{sun20253d} can answer step-level queries, but they largely frame guidance as retrieval and generation rather than as a closed-loop, state-aware process. As a result, they provide limited support for enforcing step order, detecting invalid transitions, or recovering from user deviations.

Structured task representations, including graph-based models and hierarchical planners, have improved control in robotic manipulation, but these approaches are typically designed for autonomous robots in controlled environments rather than human-facing assistance on edge devices. MetaAgent\cite{zhang2025metaagent} introduces FSM-based task planning through a generated multi-agent system, but offers limited support for runtime recovery and human feedback. AutoRT\cite{ahn2024autortembodiedfoundationmodels} uses a “robot constitution” to filter unsafe actions, though mainly as a pre-execution safety mechanism rather than a continuous runtime governance layer. Qin et al.\cite{qin2026harnessing} similarly separate monitoring and error handling from agent cognition, but do not demonstrate this design in a real-world procedural guidance setting.

Additionally, deployment of LLM-based Agentic assistant on edge and resource constraint devices is much more challenging due to their high computational demands and memory requirements\cite{tuli2025IMWUT}. There are some recent works on efficient LLM deployment on Edge hardwares\cite{shen2025mobisys} but none of them showed how a full end-to-end multi-LLMs base agentic system would deployed and seamlessly work for real-time guidance for procedural tasks. SELA\cite{tuli2025IMWUT} uses ensembling model technique for hosting only a lightweight complexity and time critically score predictors with early exit constraint to select the appropriate LLM hosted on cloud. It dooes not provide any framework to run everything on device in offline manner.

Overall, prior work leaves three important gaps: deciding when visual verification is actually needed, enforcing procedural state transitions within a conversational loop, and incorporating human feedback as a primary signal during execution. \sys~ addresses these gaps through selective perception, FSM-grounded state enforcement, and human-in-the-loop confirmation, all within a fully local edge deployment.

\begin{figure}[!htb]
    \centering
    \includegraphics[trim=60 150 50 100, clip, width=\textwidth]{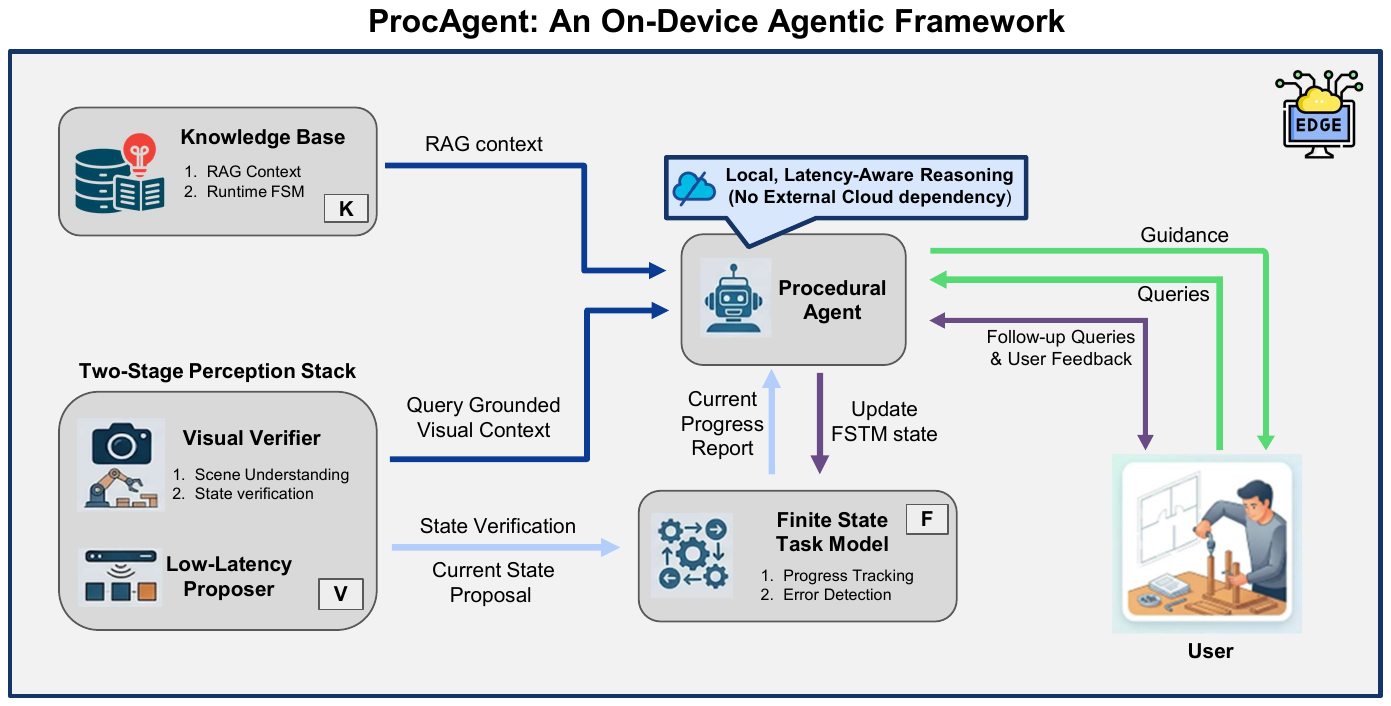}
    \caption{\textbf{System architecture of ProcAgent, an on-device agentic framework for procedural assistance.} The framework combines a \textbf{knowledge base} containing RAG context and runtime task knowledge, a \textbf{two-stage perception stack} for current-state proposal and visual verification, a \textbf{finite state task model} for progress tracking and error detection, and a central \textbf{procedural agent} for local, latency-aware reasoning. The agent integrates task knowledge, visual context, and state feedback to provide guidance to the user while processing user queries and feedback in a closed-loop interaction.}
    \label{fig:system_overview}
    \Description[]{}
    \vspace{-2mm}
\end{figure}

\section{\sys~ Framework}

\label{sec:framework}

\subsection{Overview} \label{sec:overview}
We introduce \sys, an on-device agentic framework for real-time procedural guidance during physical assembly tasks. We instantiate and evaluate \sys~on IKEA furniture assembly, a domain that exercises several demands a procedural assistant must meet: a fixed but partially ordered step structure, fine-grained physical actions that resist purely verbal description, and non-expert users who may make ordering, orientation, or completion errors. 

Figure~\ref{fig:system_overview} shows the main components of the framework and the data flow between them. The full system runs locally on edge hardware, with no external cloud dependency at runtime. Its components work together to monitor the user's progress, validate that progress against the task structure, retrieve relevant procedural knowledge, and decide when visual grounding is needed.
At the center of the framework is the \emph{procedural agent}, the LLM-based component that mediates interaction with the user. It receives user queries, decides whether a response can be generated from the current task state and retrieved procedural context, determines when visual grounding is required, produces spoken guidance, and handles follow-up when perception and task structure disagree. We call the agent’s running record of progress, errors, and recent interaction the Session Context and describe it in §\ref{sec:procagent_design}.

The \emph{knowledge base} (K in Figure~\ref{fig:system_overview}) provides the procedural context used by the agent and task model. It is constructed offline from task demonstrations and stores both RAG context for user-facing guidance and the structured task representation used to instantiate the runtime finite-state task model. We describe both the construction and the structure of the knowledge base in §\ref{sec:knowledge_base}.

The \emph{two-stage perception stack} (V in Figure~\ref{fig:system_overview}) watches the workspace. A low-latency proposer runs continuously and emits tentative current-state proposals; a higher-capacity visual verifier is invoked selectively to confirm proposed state transitions or provide query-grounded visual context. This separation lets \sys~maintain continuous awareness without running high-cost vision-language reasoning on every frame. We describe both stages in §\ref{sec:two_perception_stack}.

The \emph{finite-state task model} (F in Figure~\ref{fig:system_overview}) tracks progress and validates proposals against the legal structure of the task. Verified and legal events update the runtime FSM state and produce a current-progress report for the procedural agent; verified but illegal events are escalated to the agent for follow-up with the user. We describe its structure in §\ref{sec:FSM} and its offline construction in §\ref{sec:knowledge_base}.

Together, these components implement the propose-and-verify loop traced by the arrows in Figure~\ref{fig:system_overview}. The proposer surfaces candidate actions, the task model filters them against the current procedural state, and the visual verifier confirms only meaningful possible transitions. In parallel, user queries flow directly to the procedural agent: queries answerable from current progress and RAG context are handled without vision, while spatially grounded queries are routed to the verifier for query-grounded visual context.
This architecture reflects the central design principle of \sys: expensive visual reasoning should be invoked only when it is needed for state verification or user-facing visual grounding. The remainder of this section walks through a representative session (§\ref{sec:walkthrough}) and then details each component in turn (§\ref{sec:FSM}--§\ref{sec:procagent_design}).

\subsection{Walkthrough}
\label{sec:walkthrough}

\begin{figure}[!htb]
    \centering
    \includegraphics[trim=60 80 25 60, clip, width=\textwidth]{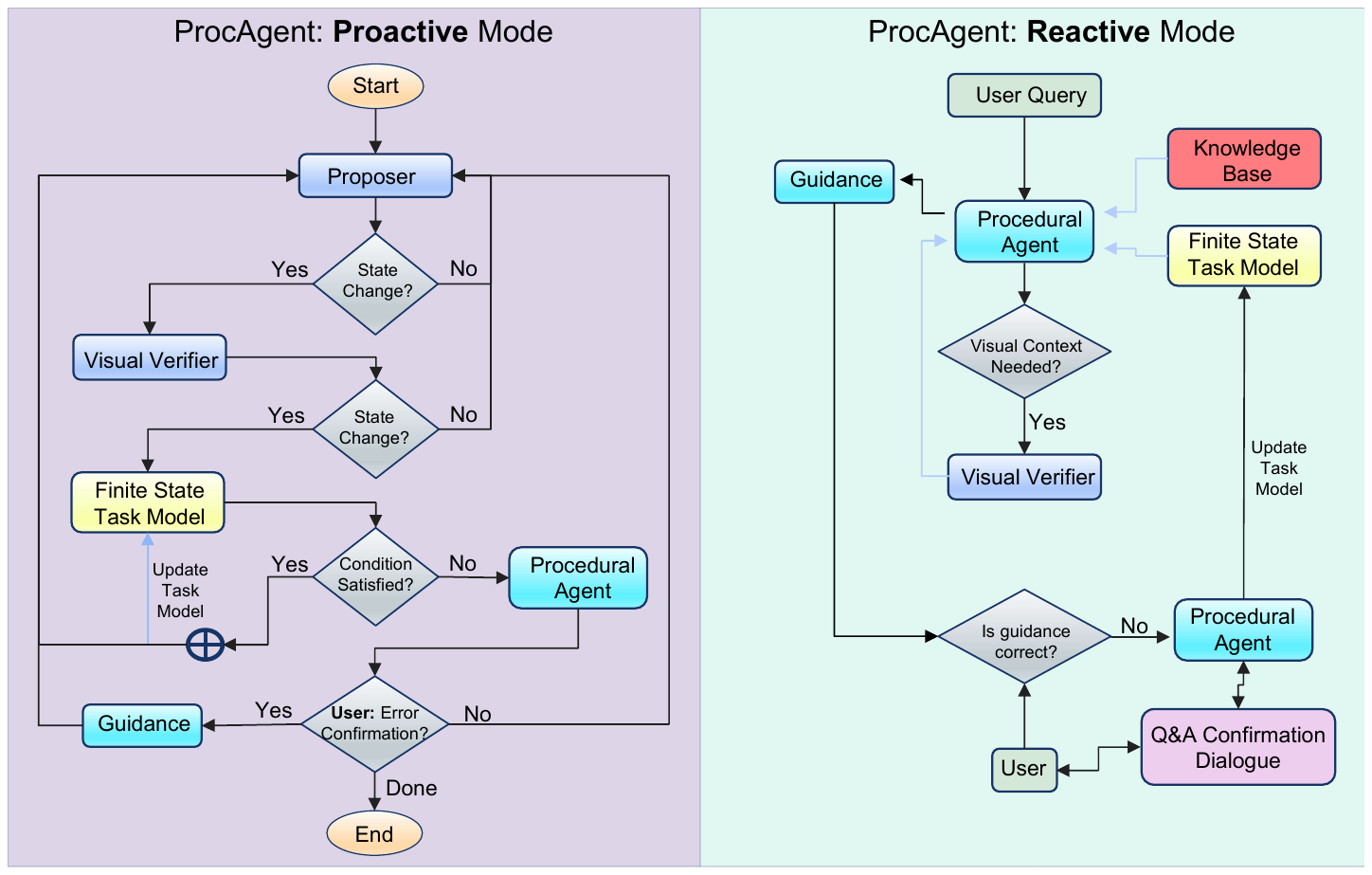}
    \caption{Flowcharts of ProcAgent’s proactive and reactive modes. In the \textbf{proactive mode} (left), the system continuously detects possible task-state changes, verifies them visually, updates the finite state task model, and issues guidance when predefined intervention conditions are satisfied. In the \textbf{reactive mode} (right), the process begins with a user query, after which the procedural agent consults the knowledge base, task model, and, when needed, visual verification to generate and confirm appropriate guidance through interactive dialogue.}
    \label{fig:walkthrough}
    \Description[]{}
    \vspace{-2mm}
\end{figure}

We illustrate how \sys~operates through two interaction modes shown in Figure~\ref{fig:walkthrough}: a \emph{proactive mode}, in which the system continuously monitors task progress and intervenes only when a likely procedural deviation is detected, and a \emph{reactive mode}, in which the user initiates a query and the agent determines whether visual grounding is needed before responding.

\paragraph{Proactive mode.}
A user begins assembling an IKEA LACK coffee table in front of a fixed workspace camera. At the start of the session, the procedural agent initializes the Session Context: the interaction history is empty and the finite-state task model is in the initial phase (\textit{Start}). The low-latency proposer begins analyzing the incoming video stream. The user picks up a leg and aligns it against the underside of the tabletop. The proposer emits a sequence of frame-level predictions, which the temporal-consistency filter smooths into a single candidate event, \textit{Align and Spin}. Because this candidate indicates a possible state change, the visual verifier is invoked to confirm whether the event is actually visible in the workspace. The verifier confirms the event. The finite-state task model then checks whether the event is legal under the current task state. Because the event satisfies the task model's guard conditions, the task state is updated. No user-facing message is produced and the monitoring loop continues. This is the common case in proactive mode: visually confirmed and procedurally valid progress is committed silently.

The same loop also handles noisy perception. If the proposer emits a candidate event because of hand motion, occlusion, or a transient visual ambiguity, but the visual verifier does not confirm the event, the candidate is suppressed. The task model is not updated, the procedural agent does not interrupt the user, and the system returns to monitoring. This suppression path prevents momentary perception errors from becoming user-facing interventions.

A different branch occurs when the workspace evidence and task structure disagree. Suppose the user attaches only one leg, then begins lifting the partially assembled table to reorient it. The proposer emits \textit{Flip Table}, and the visual verifier confirms that the user appears to be flipping the table. The finite-state task model then evaluates the event's guard and finds that it is not satisfied: the flip is only legal after all four legs have been attached. Because the event is visually confirmed but procedurally invalid, the procedural agent is invoked. Rather than immediately asserting an error or blocking progress, the agent asks the user for confirmation. If the user confirms the deviation, the agent retrieves the relevant prerequisite guidance from the knowledge base and instructs the user to complete the missing step. If the user rejects the system's interpretation, the agent suppresses the intervention, preserves the prior task state, and returns to monitoring. In this way, proactive mode treats the user as the final authority when perception and task structure disagree.

\paragraph{Reactive mode.}
The user can also initiate interaction at any time. In reactive mode, a user query is sent directly to the procedural agent, which reasons over the Session Context, the finite-state task model, and retrieved procedural context from the knowledge base. For queries that can be answered from the current task state alone, such as \textit{what is my next step?} or \textit{how many steps are left?}, the agent responds directly without invoking the visual verifier.

For visually grounded queries, the agent takes a different path. If the user asks \textit{is this leg aligned correctly?} or \textit{am I doing this right?}, the agent determines that current workspace evidence is needed and invokes the visual verifier. The verifier provides query-grounded visual context, which the agent combines with the Session Context and retrieved procedural knowledge to generate guidance. If the user indicates that the guidance is incorrect or unclear, the agent enters a brief follow-up dialogue, using the user's feedback to revise the response or defer to the user's judgment.

Together, these two modes show how \sys~uses the same components for different interaction needs. In proactive mode, perception initiates the loop: the proposer surfaces candidate events, the verifier confirms them, and the task model determines whether progress is legal. In reactive mode, the user initiates the loop: the procedural agent decides whether the query can be answered semantically or requires visual grounding. The remainder of this section develops each component in turn (§\ref{sec:FSM}--§\ref{sec:procagent_design}).

\subsection{Finite-State Task Model}
\label{sec:FSM}

The finite-state task model encodes the procedural structure of the task. It consists of a task graph, extracted offline from a task-specific corpus of demonstration videos, and a runtime state tracker that maintains the user's current position in the graph and validates candidate events proposed by the perception stack.

\subsubsection{Procedural Graph.}
The task graph is a counter-augmented finite state machine, defined by four kinds of structure. Phases partition the task into coarse stages. Events correspond to individual sub-steps and are grouped under a phase. Counters are integer variables that track how many instances of each sub-step have been completed. Guards are conditions over the counters that determine when an event may fire. The dynamics follow from these: an event fires when its guard is satisfied, increments one or more counters, and phase transitions are themselves events whose guards reference the counters. So a phase advances only when its sub-steps have completed in a configuration consistent with the next phase's requirements. The graph is extracted automatically from a corpus of demonstration videos; we describe the extraction procedure in \ref{sec:knowledge_base}.

\begin{figure}[H]
    \centering
    \includegraphics[trim= 120 300 60 160, clip, width=\textwidth]{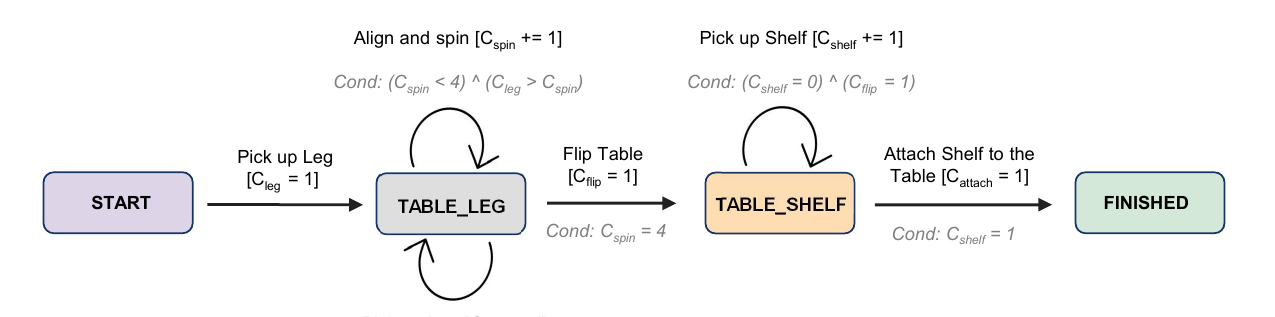}
    \caption{Counter-augmented finite-state task graph for the furniture assembly instantiation. Phases (rounded rectangles) partition the task into coarse stages; events (labeled arrows) correspond to sub-steps and update integer counters; guard conditions on events determine when each event may fire. Self-loops within \textsc{table\_leg} allow the four leg attachments to be performed in any order while still enforcing that each leg is aligned and spun after it is picked up.}
    \label{fig:task_model}
    \Description[]{}
    \vspace{-2mm}
\end{figure}

For the assembly tasks we evaluate in this paper, the task graph has four phases (\textit{Start}, \textit{Table-Leg}, \textit{Table-Shelf}, \textit{Finished}) and five events (\textit{Pick Up Leg}, \textit{Align and Spin}, \textit{Flip Table}, \textit{Pick Up Shelf}, \textit{Attach Shelf}). Within the Table-Leg phase, \textit{Pick Up Leg} and \textit{Align and Spin} self-loop with guards on two counters ($C_\text{leg}$ and $C_\text{spin}$) that enforce two properties: each leg must be aligned and spun after it is picked up (so $C_\text{leg}$ and $C_\text{spin}$ stay synchronized to within one), and no more than four legs can be processed ($C_\text{leg} \leq 4$, $C_\text{spin} \leq 4$). The \textit{Flip Table} event transitions the phase to Table-Shelf, with the guard that all four legs have been completed before the flip is permitted. Figure~\ref{fig:task_model} shows the full task graph for this instantiation.

The counter-augmented structure is what lets the task graph remain compact in the presence of parallelism. A representation that enumerated every ordering of the four legs as a separate path would produce $4!=24$ paths through the Table-Leg phase alone; the counter representation collapses these into a single phase with self-loops, at the cost of requiring the counters to be tracked explicitly at runtime. The graph is fixed at runtime; user feedback during a session updates the active phase and the counters but does not modify the graph itself.

\subsubsection{State Tracking.}
At runtime, the task state consists of the current phase and the current values of the counters associated with that phase. The state tracker updates the task state when a candidate event proposed by the perception stack has been verified (§\ref{sec:two_perception_stack}) and the corresponding event's guard is satisfied.

The task model is consulted twice per candidate. On receipt of a candidate, it identifies the matching event out of the current phase; a candidate that matches no event is treated as a no-op and discarded. After the candidate has been confirmed by the visual verifier (§\ref{sec:visual_verifier}), the task model evaluates the event's full guard. A guard-satisfying event fires: the counters are incremented and, if the event is a phase-transition event, the phase is advanced. A guard-violating event is escalated to the procedural agent (§\ref{sec:procagent_design}).

The task model also serves as a read interface when the procedural agent is answering user queries about the session. Queries such as \textit{what is my next step?} or \textit{how many steps are left?} are answered from the events whose guards are satisfiable under the current (phase, counter) state. This query path is separate from the state-tracking path: reading the current satisfiable events does not advance the task state.

The task graph and the verification VLM catch different kinds of errors. The graph rules out events whose guards are not satisfied in the current state, which captures procedurally impossible transitions such as attempting to flip the table before all four legs are attached. The VLM rules out events whose guards would be satisfied but that the workspace does not actually show, which captures perception errors and transient misclassifications. Neither mechanism catches the other's failure mode: the graph does not detect a visually absent event that happens to be guard-satisfiable, and the VLM does not detect a visually present event that happens to violate its guard. The two checks are applied in sequence so that a committed state update has cleared both.

\subsection{Two-Stage Perception Stack}
\label{sec:two_perception_stack}

The perception stack maintains continuous awareness of the workspace and surfaces evidence for state-tracking decisions made by the task model. It consists of a lightweight proposer that processes incoming video frames and emits candidate task events, and a higher-capacity visual verifier that confirms or disconfirms each candidate before it is committed. A key design choice in \sys is to treat the visual verifier as an on-demand visual tool rather than an always-on perception stream. The proposer runs continuously and is cheap enough to invoke at video rate, while the visual verifier is invoked only when a candidate event requires confirmation or when the procedural agent determines that a user query depends on the current workspace. This division lets the stack maintain low-latency awareness without applying high-capacity visual reasoning to every frame.

\subsubsection{Proposer.}
\label{sec:proposer}

The proposer runs in the background and maps each incoming video frame to a predicted task event in a shared image-text embedding space. We instantiate the proposer with a CLIP model fine-tuned on task-specific examples (see Section \ref{sec:proposer_finetune} for the training details). For each frame, the proposer computes the frame's embedding and its nearest task-event description under the fine-tuned similarity function, producing a per-frame event prediction.

We do not treat the proposer's per-frame output as ground truth. Fine-grained procedural tasks are precisely the setting where embedding-based classifiers are most susceptible to noise: components look similar across events, hands frequently occlude salient parts, and lighting conditions shift during a session. Treating every frame-level prediction as a state-tracking signal would generate a continuous stream of false positives. Instead, the proposer aggregates per-frame predictions into candidate events using a sliding-window vote. A buffer of the most recent frame predictions is maintained, and at fixed intervals the proposer emits the majority prediction across the buffer as a single candidate. We call this ``proposer buffer.'' The window absorbs transient flickers and brief occlusions rather than propagating them to the task model.

The proposer is indifferent to whether a given candidate represents a change of state or a continuation of the current state: it emits candidates that match the current task state as well as candidates that do not. The task model (§\ref{sec:FSM}) is responsible for distinguishing claimed transitions from no-op predictions, and discards candidates that match the current state without invoking the visual verifier. The proposer's role is to produce stable candidates cheaply, not to reason about whether a candidate advances the task.

\subsubsection{Visual verifier.}
\label{sec:visual_verifier}
The visual verifier is a higher-capacity vision-language model that the stack invokes on two paths. The first is state-tracking verification, triggered by the task model whenever the proposer emits a candidate that corresponds to a claimed task event. The second is query grounding, triggered by the procedural agent whenever a user query requires visual evidence from the current workspace (§\ref{sec:procagent_design}). The two paths share the same underlying model but use different prompts and consume the model's output differently.

For state-tracking verification, the prompt is generated from the specific candidate event under evaluation. The visual verifier is shown a small set of representative frames from the proposer buffer, together with the event description, and asked whether the event is actually occurring in the workspace, returning a yes/no verification. When the verifier returns \textit{yes}, the task model proceeds to the guard check described in §\ref{sec:FSM}; when it returns \textit{no}, the candidate is suppressed and the session continues without interrupting the user. The targeted per-event prompt is important: asking a general-purpose visual question such as \textit{what is happening in the workspace?} produces free-form output that is harder to map back to a specific event and more susceptible to distraction by unrelated scene content.

The task model also limits unnecessary verification by filtering proposer outputs before invoking the visual verifier. Candidate events that do not correspond to a possible event from the current task state are discarded, and candidates that match the current state are treated as no-ops. Thus, high-capacity verification is reserved for proposer outputs that represent potential task progress and require visual confirmation before they can be committed.

For query grounding, the prompt is constructed by the procedural agent from the user's query and the current Session Context, so that the model's attention is focused on the task-relevant properties the query is asking about. The model's output in this path is free-form natural language, because the user's question is not a yes/no verification. The agent consumes the output as context for response generation rather than as a gating signal. The visual verifier fires only when one of these two paths invokes it; frames without an emitted candidate or an active user query are not processed.

\subsection{Procedural Agent Design}
\label{sec:procagent_design}
The procedural agent is the component the user interacts with and the component that mediates between perception and the task model when user-facing action is required. We instantiate the agent with a pretrained large language model that, on each invocation, receives the current Session Context and a description of available actions (retrieve from the knowledge base, invoke the verification VLM, produce a response) and selects among them. The agent has three responsibilities. First, it maintains the Session Context, a structured record of the user's progress through the current session. Second, it mediates error handling when the task model reports that a confirmed event violates its guard, conducting a confirmation dialogue with the user and generating recovery guidance when appropriate. Third, it handles user-initiated queries, including the decision of whether a query can be answered from the Session Context alone or whether the verification VLM must be invoked to ground the response. The agent does not sit in the state-verification path for routine events: when the task model and VLM agree that an event is visually present and guard-satisfying, the state update is committed without any involvement from the agent.

\subsubsection{Session Context.}
The Session Context is the agent's working memory across a session. It records the interaction history between user and agent, the current task state (phase and counter values), and a log of errors and corrections that have occurred so far. The Session Context is updated at three points: whenever a state transition is committed by the task model, whenever the user and agent exchange an utterance, and whenever a confirmation dialogue resolves to a definite outcome. The agent consults the Session Context before every response it produces, so that its output is conditioned on the user's current progress rather than on a generic reading of the task.

The choice to maintain a persistent, structured session record changes how the agent answers questions that depend on history. A query such as \textit{what went wrong?} has no fixed answer; the answer depends on which error the user just encountered, which step they are in, and what the agent most recently said. The Session Context makes these dependencies explicit and addressable. Responses that reference earlier events in the session, such as ``you skipped attaching the third leg before flipping the table,'' are generated from the error log and the current task state in the Session Context, not recovered from conversation history alone.

\subsubsection{Perception-Loop Mediation, Error Handling and User Feedback.}
The agent's role in the perception loop is limited to handling escalations. Most candidates emitted by the proposer are resolved upstream: the task model discards candidates that do not claim a transition, the verification VLM suppresses those that are not visually present, and guard-satisfying events are committed without agent involvement. The agent is invoked only in the remaining case, when the VLM confirms a candidate but its guard is not satisfied in the current state. The workspace evidence and the procedural structure disagree, and resolving the disagreement requires interaction with the user.

The agent handles escalation through a confirmation dialogue rather than autonomous action. It informs the user of the apparent deviation, grounded in the specific event whose guard was violated, and asks the user to confirm or reject the observation. Two branches follow. If the user confirms, the agent retrieves guidance for the prerequisite step from the knowledge base and generates a recovery instruction; the Session Context records the error and the task state is held until the user returns to a valid configuration. If the user rejects, the agent suppresses the intervention, retains the prior task state, and does not block further progress. Subsequent candidates are still evaluated against the task model, so a single rejected confirmation does not disable error detection for the remainder of the session. Additionally, user can also correct the agent's response. \sys~features a user-initiated correction mechanism to resolve factual discrepancies in the agent's output. Upon detecting an error in the response, the user can trigger a synchronization event that shifts the agent into a structured Q\&A confirmation dialogue. In this mode, the agent systematically verifies the execution status of all prior milestones, allowing it to reconcile its internal session context and update its task graph based on validated user feedback.

This design yields to the user on disagreement rather than blocking progress. The rationale is practical: the VLM can misclassify, the task graph can encode a procedural expectation that the user has reasonable grounds to deviate from, and the cost of a false intervention is higher than the cost of a missed one in a voluntary-use setting. We discuss the tradeoffs of this choice in §\ref{sec:ablation}.

\subsubsection{Query Handling and Response Generation.}
User queries can arrive at any point in the session and are handled on a separate path from the perception loop. The central decision is whether a query can be answered from the Session Context and the task graph alone, or whether it requires visual grounding in the current workspace. We refer to this decision as \textbf{Reason-Before-Perception}: before any vision resources are invoked to answer a query, the agent reasons over the query text and the Session Context to decide whether visual evidence is actually necessary for the response.

The routing decision is made by the agent's LLM. The LLM receives the user's query, the current task state, and a compressed view of the Session Context, and outputs a structured decision indicating whether the verification VLM should be invoked for this query. This is a distinct mechanism from the state-tracking VLM invocation: in the perception loop the VLM fires on every claimed event emitted by the proposer, whereas here the VLM is invoked only when the routing decision calls for it. Queries such as \textit{what is my next step?} or \textit{how many steps are left?} typically route without vision, because their answers are fully determined by the current task state and the currently satisfiable events in the task graph. Queries such as \textit{am I doing this correctly?} or \textit{is this leg aligned?} typically route with vision, because their answers depend on properties of the current workspace that the Session Context does not encode.

When the routing decision calls for vision, the agent invokes the verification VLM with a prompt constructed from the query and the current Session Context, so that the model's output addresses the step-specific properties the query is asking about rather than the scene as a whole. The response to the user is then generated by the agent's LLM from three inputs: the query, the current Session Context, and context retrieved from the knowledge base. The retrieved context consists of step-level records for the current and adjacent task-graph events, drawn from the same corpus of demonstration videos from which the task graph is mined (see §\ref{sec:knowledge_base} for the construction). For vision-routed queries, the VLM's output is added to the prompt as an additional input describing the current workspace. This generation path is the same for both routes; the only difference is whether the VLM's workspace description is present in the prompt.

The Session Context that the agent provides as input to the response-generation LLM is not the full session record. Long interaction histories are compressed into a summary that preserves the current task state, the recent error log, and the last few user-agent turns. This compression is necessary because the LLM context window is bounded, and it is consistent with the routing LLM's input, so that both the routing decision and the response generation are conditioned on the same view of the session.

\section{Implementation: Task Setup and On-Device Execution}

Section~\ref{sec:framework} described \sys as a task-structured agentic framework composed of a procedural agent, a finite-state task model, and a two-stage perception stack. In this section, we describe how these abstract components are instantiated in our implementation and executed on the edge device. The implementation is organized around a separation between \textbf{task-specific knowledge} and \textbf{task-independent runtime logic}: the Knowledge Base and FSM configuration change across tasks, while the agent controller, perception stack, memory budgeting strategy, and inference pipeline remain fixed.

\subsection{Implementation Overview and Model Roles}

To instantiate \sys for a new procedural task, the developer provides a task-specific corpus of demonstration videos and a fixed camera setup for the target workspace. \sys processes the demonstrations offline to extract atomic steps, validate them across models, and generate a structured Knowledge Base. This Knowledge Base is then integrated with the generic runtime FSM engine to produce a task-specific state model. At deployment time, the resulting task-specific state model is loaded into the on-device runtime, where the lightweight proposer, visual verifier, procedural agent, and response-generation pipeline execute locally on the edge device.

Table~\ref{tab:model_role} summarizes how each framework component is instantiated in our implementation. GPT-OSS-20B serves as the procedural agent for query routing, Session Context management, tool invocation, and response generation. The two-stage perception stack consists of a  proposer implemented with fine-tuned CLIP and a verification model implemented with Qwen2.5-VL-3B. The proposer continuously processes incoming video frames and emits low-cost candidate task events, while the verifier is invoked only when visual grounding is required, either to confirm a proposed state transition or to answer a workspace-dependent user query. The finite-state task model is implemented as a counter-augmented FSM that enforces legal task order, validates event guards, and tracks the current phase and counter state. The Knowledge Base is constructed offline using Gemini and GPT-4o to extract, validate, and structure task knowledge from demonstration videos before deployment.

\begin{table}[t]

\centering

\caption{Model roles in the \sys implementation.}
\label{tab:model_role}

\begin{tabular}{p{0.22\linewidth}p{0.24\linewidth}p{0.44\linewidth}}

\toprule[2pt]

 System role & Model / component & Purpose \\

\midrule

Procedural Agent & GPT-OSS-20B & Routes user queries, manages Session Context, invokes tools, and generates responses. \\

Visual Verifier & Qwen2.5-VL-3B & Confirms proposed state transitions and grounds visually dependent user queries. \\

Proposer & Fine-tuned CLIP & Continuously monitors frames and proposes candidate task events. \\

Finite-State Task Model & Counter-augmented FSM & Enforces task order, validates guards, and tracks phase/counter state. \\

Knowledge Base Construction & Gemini + GPT-4o & Extracts, validates, and structures task knowledge offline from demonstration videos. \\


\bottomrule[2pt]

\end{tabular}

\end{table}

This mapping fixes the implementation boundary for the rest of the section: task-specific knowledge is produced offline and loaded into the FSM and Knowledge Base, while the same on-device proposer, visual verifier, and procedural agent are reused across task instances.

\subsection{Knowledge Base Construction and Integration}
\label{sec:knowledge_base}

\begin{figure}[!htb]
    \centering
    \includegraphics[trim= 30 130 110 110, clip, width=\textwidth]{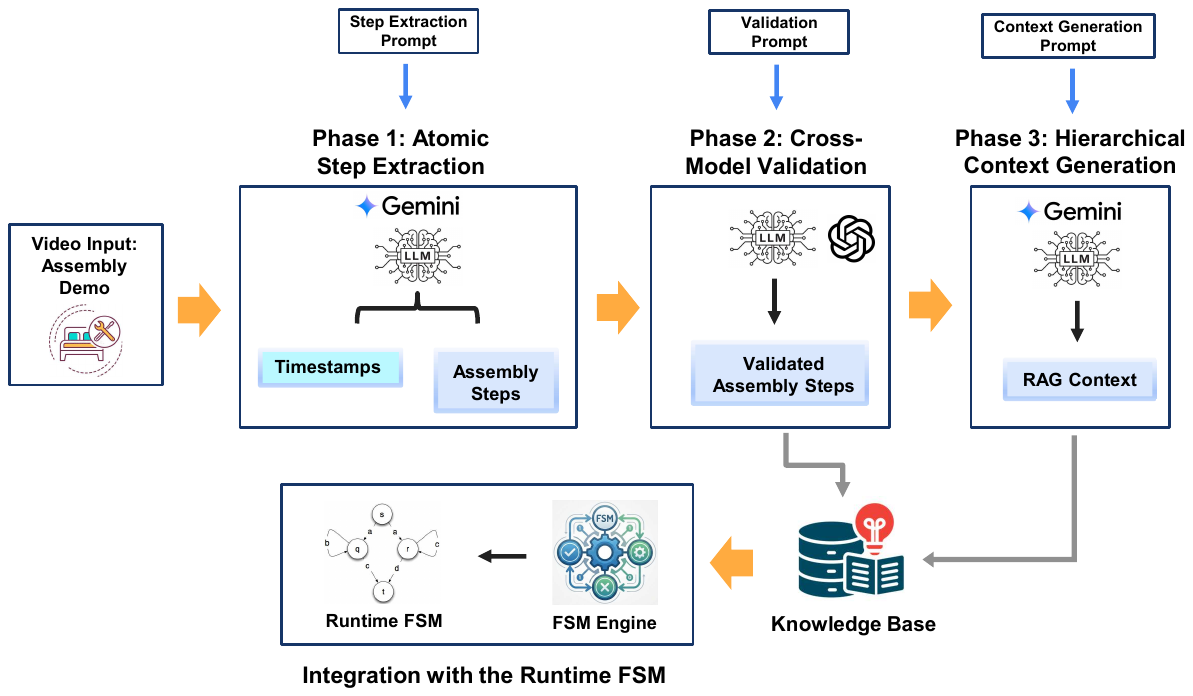}
    \caption{Offline knowledge base construction pipeline. Demonstration videos are processed by a three-stage pipeline (atomic step extraction, cross-model validation,  hierarchical context generation) to produce the Knowledge Base. A deterministic integration step configures the generic FSM engine for the new instance, producing the runtime-ready finite-state task model.}
    \label{fig:knowledge_base_construction}
    \Description[]{}
    \vspace{-2mm}
\end{figure}

The Knowledge Base is constructed once per task instance from a corpus of demonstration videos and reused across all subsequent sessions for that instance. Construction is performed offline using cloud language models; no cloud calls are made at inference time. Figure~\ref{fig:knowledge_base_construction} summarizes the offline process: a three-stage pipeline produces a structured Knowledge Base, and a separate integration step reads the Knowledge Base to instantiate the runtime finite-state task model (§\ref{sec:FSM}) for that instance.

The pipeline has three stages: atomic step extraction, cross-model validation, and hierarchical context generation. A fourth integration step then reads the Knowledge Base produced by the pipeline and configures the runtime FSM.


\subsubsection{Atomic Step Extraction}
The first stage uses Gemini's long-context video reasoning capability to perform dense temporal analysis over the demonstration videos for the instance. The model extracts atomic steps, the smallest discrete units of action (such as ``pick up leg'' or ``align and spin''), each paired with start and end timestamps. The extraction prompt enforces a low-level granularity heuristic so that compound or ambiguous actions are decomposed into sub-tasks small enough to be independently verifiable by the perception stack at runtime. The prompt is included in the Appendix \ref{appendix:step_extract_prompt}.


\subsubsection{Cross-Model Validation}
To mitigate the risk of single-model extraction bias, a second model (GPT-4o) acts as an independent validator over the Gemini-extracted steps. The validator audits each step for logical consistency, checks for temporal overlaps between adjacent steps, and flags any step whose description does not align with its assigned timestamp window.

\subsubsection{Hierarchical Context Generation}
The validated steps are then organized by Gemini into a hierarchical context (structured task description) that the runtime system can consume. The description has three components: a sequence of high-level steps in their canonical execution order, a set of completion rules associating each step with a numeric completion count (e.g., pick up leg: ``4 times''), and a hierarchical narrative that records inter-step dependencies (e.g., ``after attaching four legs, ``flip the table''). Steps with repeated sub-actions, such as ``align and spin'' performed once per leg, are explicitly assigned numeric repetition counts so that downstream components can ground state transitions in countable success criteria rather than soft model confidence.

This structured task description serves two downstream uses. The integration step reads it to configure the runtime FSM. The procedural agent (§\ref{sec:procagent_design}) retrieves from it as the RAG context that grounds response generation during live sessions: when the agent answers a user query or generates recovery guidance after an error, it draws on this description to anchor the response in the specific procedural structure of the current task instance.

\subsubsection{Integration with the Runtime FSM}
The finite-state task model described in §\ref{sec:FSM} is implemented as a generic FSM engine, hand-authored once and shared across all task instances. To prepare the engine for a new instance, a deterministic integration step reads the structured task description from the Knowledge Base and configures the engine: each step in the canonical sequence becomes an event in the FSM, each completion rule becomes a counter target, and the inter-step dependencies become guard conditions on event firing. The phase structure (e.g., \textit{Start}, \textit{Table-Leg}, \textit{Table-Shelf}, \textit{Finished}) is determined by hand-coded rules in the integration step, applied uniformly across instances. The output is a runtime-ready FSM that the engine instantiates at startup. The Knowledge Base also remains available to the procedural agent for retrieval during user queries.


\subsection{Memory Orchestration \& Resource Budgeting}

Running \sys entirely on the edge device requires coordinating multiple models under a shared memory budget. Our implementation targets the \textbf{NVIDIA Jetson AGX Orin}, whose unified memory architecture exposes a single 64GB memory pool shared by the CPU and GPU. This design simplifies data movement between components but also makes memory contention a central runtime constraint: the procedural agent, visual verifier, and proposer all draw from the same memory pool, along with their associated runtime allocations such as KV caches, vision-encoder activations, frame buffers, audio I/O, and operating-system overhead. We therefore treat model placement, quantization, and tool invocation as part of the system design rather than as independent implementation details.

We allocate memory according to the role each model plays in the agentic pipeline. The procedural agent is the largest and most important language-reasoning component, so GPT-OSS-20B is quantized to Q5\_K\_M to reduce its footprint while preserving response quality. The visual verifier, implemented with Qwen2.5-VL-3B, is also quantized to Q5\_K\_M, since it is used for visually grounded verification and query answering but is not executed continuously. The proposer, implemented with fine-tuned CLIP, remains small enough to run continuously for frame-level monitoring. This allocation leaves headroom for dynamic KV-cache growth, vision-encoder activations, frame buffers, audio I/O, and operating-system overhead during interactive sessions.

In our implementation, state-tracking verification uses three representative frames from the proposer buffer instead of a single frame or the full window. We sample the middle, third-quarter, and final frames of the candidate window, giving the visual verifier short temporal context because assembly actions unfold over nearby frames and may appear as preparation, mid-action, or completion. The visual verifier receives these frames with the candidate event description and returns a binary decision indicating whether the event is present in the workspace.

Before verifier invocation, the task model filters no-op candidates and candidates that do not correspond to a possible event from the current task state. Together, quantized model deployment, three-frame prompting, selective verifier invocation, and task-model filtering allow \sys to run a multi-model agentic pipeline within the Jetson memory envelope. Three-frame prompting bounds the cost of each verification call, while task-model filtering reduces the number of calls. As a result, \sys preserves continuous low-cost perception while reserving high-cost reasoning for state verification, error handling, and visually grounded user assistance. This avoids the sustained GPU compute pressure and memory-bandwidth contention of an always-on VLM stream, helping \sys remain responsive under edge-device constraints.

\subsection{Inference Pipeline}

The runtime inference pipeline is implemented using \emph{llama.cpp}, an optimized inference runtime that supports quantized model execution on ARM-based edge hardware. We serve the quantized procedural agent and visual verifier through the \emph{llama.cpp} server interface with parameters tuned to the Jetson's memory and compute constraints. This allows the language and vision-language components to run locally without cloud inference while keeping their memory footprint compatible with the shared 64GB unified memory budget.

At the start of a session, the task-specific FSM configuration and Knowledge Base are loaded into the runtime. The proposer then begins processing incoming video frames continuously and emits temporally smoothed candidate events. The inference loop handles two input streams: proposer-generated candidates from the perception path and user-initiated queries from the interaction path. Candidate events are filtered by the task model and, when needed, routed to the visual verifier before any state update is committed. User queries are routed by the procedural agent either to direct response generation from the Session Context and Knowledge Base or to visually grounded response generation using the visual verifier.

The pipeline is structured as a non-blocking loop so that continuous perception, visual verifier, and response generation do not unnecessarily stall one another. The proposer continues monitoring the workspace while the agent handles user queries, and the visual verifier is invoked only for bounded verification calls or workspace-dependent queries. This implementation preserves the framework behavior described in Section~\ref{sec:framework} while making the runtime practical under the Jetson's memory, compute, and latency constraints.



\section{On-Device System Evaluation}

This section evaluates \sys as an on-device procedural guidance system. We first describe the edge hardware, assembly dataset, system configurations, and evaluation metrics. We then evaluate the low-latency proposer, the proposer buffer and visual-verifier frame-selection strategy, the contribution of each system component, and the impact of procedural-agent backbone choice. Finally, we analyze the resource and latency trade-offs of running the complete pipeline on the Jetson AGX Orin.

\subsection{Experimental Setup}
This subsection describes the setup used for the on-device evaluation. We summarize the hardware, assembly dataset, system configurations, evaluation metrics, and user-query bank used in the experiments.

\subsubsection{Hardware}
\label{ssubsec:hardware}
All experiments are conducted entirely on a single NVIDIA Jetson AGX Orin~\cite{nvidia2021jetsonagxorin} (64 GB unified memory, Ampere GPU, 12-core ARM Cortex-A78AE) running in MAXN power mode, with no cloud inference during evaluation. The workspace is captured by a Logitech C920s Pro camera~\cite{logitechc920s} (1080p, 30,fps); the agent's responses are broadcast through an EMEET Luna Conference Speaker~\cite{emeet_luna}, and user queries are captured by a directional microphone.


\subsubsection{Evaluation Dataset} 

We evaluate the system on a curated subset of the IKEA Assembly Dataset\cite{ben2021ikea}. From the dataset's 371 videos, we evaluate on 15 covering three furniture types (LACK Coffee Table, LACK Side Table, LACK TV Bench), with five demonstrations per type. The subset is constructed under three criteria. First, each video must exceed 100 seconds in duration, ensuring that the demonstration includes the full assembly sequence rather than a partial or aborted attempt. Second, we exclude videos in which the user performs multiple sub-steps concurrently, because our task model commits to a single active step at any given time and the sequence-adherence metric is defined with respect to that assumption; we discuss concurrent execution as future work in Section [limitations]. Third, the retained videos span multiple users, lighting conditions, and table colors, to capture variation across realistic recording conditions within the dataset. The list of selected videos is provided in Appendix \ref{appendix:video_data_list}.

\subsubsection{Proposer Fine-Tuning} 
\label{sec:proposer_finetune}
The proposer (a fine-tuned CLIP, see §\ref{sec:proposer}) is fine-tuned on a separate set of 29 demonstration videos drawn from the same IKEA Assembly Dataset. The fine-tuning videos cover the only one type of furniture but feature different users; no video used for fine-tuning appears in the evaluation set. No other component of the pipeline is trained on either set: the visual verifier (Qwen2.5-VL-3B) and procedural agent (GPT-OSS-20B) are used as pretrained off-the-shelf models, and the Knowledge Base for each furniture type is constructed offline from different demonstration videos than the evaluation dataset.


\subsubsection{Baseline} 
\label{eval_baseline}
We evaluate \sys through two complementary axes: an ablation across system configurations and a comparison across agent backbones.


\noindent\textbf{System Ablation.} 
The procedural agent and the proposer are enabled in all configurations of the system; we vary the visual verifier's two roles, the finite-state task model, the Knowledge Base retrieval, and the user-confirmation dialogue. Table~\ref{tab:ablation_results} summarizes the configurations. Configuration A includes the procedural agent and the proposer alone; the agent generates responses from the proposer's raw state proposals and the Session Context, with no verification, state enforcement, knowledge retrieval, or user confirmation. Configuration B adds the visual verifier in its state-tracking role (§\ref{sec:two_perception_stack}). Configuration C adds the finite-state task model on top of B. Configuration D adds Knowledge Base retrieval to B without the task model, isolating retrieval's contribution against state enforcement. Configuration E enables both the task model and retrieval. Configuration F adds the visual verifier in its query-grounding role. The full \sys configuration adds the user-confirmation dialogue. The pairwise comparisons isolate the marginal contribution of each component. A versus B measures the verifier's effect on state tracking. B versus C measures the task model's effect under verifier-only operation. B versus D measures retrieval's effect under verifier-only operation. E versus F measures the contribution of query-grounded visual reasoning. F versus full \sys measures the contribution of user confirmation.

We also evaluate a separate \textit{Single VLM} baseline, reported in its Table \ref{tab:one_vlm}, in which a vision-language model answers the user’s query directly from the five most recent camera frames without explicit state tracking, retrieval, verification, or user confirmation. We treat this as an external baseline rather than an ablation configuration, since it replaces the full \sys architecture with a one-pass alternative.

We additionally tested a configuration in which the proposer was disabled and the visual verifier was invoked on every incoming frame, taking on the proposer's role of generating per-frame state predictions. We do not report this configuration in Table~\ref{tab:ablation_results} because per-frame VLM invocation is not deployable in real-time settings: a single assembly session of approximately three minutes took roughly 100 minutes to complete in this configuration, more than an order of magnitude beyond interactive latency. The measurement is consistent with the design rationale for the propose-and-verify architecture (§\ref{sec:two_perception_stack}): per-frame verification is computationally infeasible on edge hardware, which is the gap the proposer is designed to fill. We retain this measurement as a runtime calibration of the per-frame vision-language model cost rather than as an ablation row.

For the user-confirmation evaluation, we developed a rule-based simulator that activates only when the system makes a sequencing error; this reflects the intended runtime role of confirmation as a corrective signal rather than a general-purpose interaction channel.

\noindent\textbf{Agent-Backbone Comparison.} 
We benchmark three quantized open-source language models as the procedural agent: Phi-3-Mini-4k-Instruct (3.8B), LLaMA-3.1-Instruct (8B), and GPT-OSS-20B (20B). All three are served through llama.cpp at $Q5\_K\_M$ quantization for LLaMA-3.1-Instruct (8B), and GPT-OSS-20B (20B) and $Q4\_K\_M$ quantization for Mini-4k-Instruct (3.8B) as we could not find open source $Q5\_K\_M$ version of Mini-4k-Instruct (3.8B). Each backbone is evaluated under the full \sys configuration to isolate the effect of agent capacity from the rest of the pipeline. The default \sys uses GPT-OSS-20B; the alternatives are evaluated for their accuracy-latency tradeoff in §5.2.3.

\subsubsection{Metrics}
We evaluate \sys along three groups of metrics: decision accuracy, response latency, and response quality. Each metric is defined operationally below. Where a metric depends on which components a configuration includes, the relevant configuration behavior is noted in the corresponding definition.

\subsubsection*{Decision Accuracy}
These three metrics evaluate discrete decisions the system makes during a session, comparing each decision against ground-truth annotations: identifying the user's current state, classifying a candidate event as consistent or inconsistent with legal task structure, and routing a user query to the appropriate response path.

\textbf{State Accuracy} measures how often the system correctly identifies the user's current assembly state. It is computed at the candidate-event level: each time the proposer emits a candidate (after the sliding-window vote), we compare the system's final predicted candidate to the ground-truth candidate annotated for the corresponding video segment. In configurations with the visual verifier and the finite-state task model, the predicted candidate is the candidate that the FSM commits after verifier confirmation; in Version A, it is the candidate emitted directly by the proposer. State Accuracy is the percentage of candidate emissions whose final predicted candidate matches the ground truth. Ground-truth annotations were collected from the IKEA Assembly Dataset's\cite{ben2021ikea} gt\_segments label.

\textbf{Sequence Adherence} measures how reliably the system enforces legal procedural order. Over a session, candidate events may arise that are inconsistent with the legal task structure, for example a flip-table event before all four legs have been attached. The denominator is the set of candidates that reach the state-update decision in each configuration (proposer-emitted in Version A, verifier-confirmed otherwise); the numerator is the count of those candidates that the system correctly classified as consistent or inconsistent with the legal task structure. A score of 100\% means every reaching candidate was correctly classified; a lower score reflects either an illegal candidate that was committed or a legal candidate that was incorrectly rejected. Because the denominator differs across configurations, the metric is not a unit-comparable quantity across rows of Table~\ref{tab:ablation_results}; we report it as a per-configuration measure of procedural-control reliability.

\textbf{Routing F1} measures the accuracy of the procedural agent's routing decision on the user-query path: whether a given user query is routed to direct response generation from the Session Context and Knowledge Base, or to a verifier-grounded response (§3.5.3). A true positive is a routing decision that invokes the visual verifier when visual context was genuinely necessary; a false positive is a verifier invocation when language context alone was sufficient; a false negative is a query that required visual grounding but was answered without it. Ground-truth labels for the 20 unique query templates from the question bank (§5.1.5) were produced by a researcher's independent annotation. Routing F1 covers only the user-query path. Verifier invocation on the perception path is rule-based: every confirmed candidate state change triggers a verifier call by construction, so this path does not involve a model decision and is not evaluated by this metric.

\subsubsection*{Response Latency}
Average Response Time measures the latency the user directly experiences when querying the agent, while Average Prediction Time measures the latency between a state change in the workspace and the system's recognition of it. Together, the two metrics characterize the responsiveness of the system on its user-initiated and system-initiated paths.

\textbf{Average Response Time} measures the latency of the agent's reactive responses to user queries, from query receipt to the start of the agent's full grounded response. It excludes the proactive acknowledgment ("hold on, please") that the deployed system uses to signal that a longer response is being prepared; that acknowledgment is part of the proactive interaction design and is discussed in §6. Within the metric, response time varies by query route: queries answered from the Session Context and Knowledge Base alone are bounded by the procedural agent's generation time, while queries routed to the visual verifier additionally include the verifier's confirmation time.

\textbf{Average Prediction Time} measures the total time required to produce one verified candidate event, from the start of the proposer's window to the FSM's accept-or-reject decision. This includes the proposer's processing of its sliding window (CLIP inference over 15 frames at stride 15) and, when applicable, the verifier's confirmation time. The verifier component is zero in Version A and on candidates that the task model filters as no-ops. The metric is averaged over all candidate emissions in the evaluation set.

\subsubsection*{Response Quality}
The three response-quality metrics distinguish three different ways an agent's response can fail. Factual Consistency penalizes responses that introduce information not supported by the supplied evidence (sins of commission). Correctness penalizes responses that distort or contradict the supplied evidence (sins of distortion). Helpfulness penalizes responses that, even when free of the above failures, do not move the user forward on the task (failures of actionability). The three rubrics are intended to be evaluated independently, even when they correlate empirically.

\textbf{Factual Consistency Score} (0--5) measures whether the agent's response is consistent with the Knowledge Base and the verified current state, in the sense that every claim in the response can be traced to one of these sources. A score of 5 indicates a response in which all claims are supported by the supplied evidence; lower scores reflect increasing degrees of unsupported or invented content, down to 0 for a response entirely unrelated to the task. A response that is factually true in the world but contains claims absent from the Knowledge Base is treated as inconsistent under this metric, since the metric evaluates fidelity to the supplied evidence rather than to ground truth about the world~\cite{kryscinski1910evaluating}.

\textbf{Correctness Score} (0--5) measures whether the agent's response correctly reflects the requirements of the current step as encoded in the Knowledge Base. Where Factual Consistency penalizes invented claims, Correctness penalizes claims that contradict or distort the supplied evidence: for example, telling the user to attach the shelf when the Knowledge Base prescribes flipping the table at the current step. A score of 5 indicates a response that correctly characterizes the current step and its requirements; lower scores reflect increasing degrees of factual error, down to 0 for a response that is completely wrong.

\textbf{Helpfulness Score} (0--5) measures whether the agent's response effectively helps the user make progress on the task. A response can be factually consistent and correct but still unhelpful if it is vague, incomplete, or fails to specify the action the user should take next. A score of 5 indicates a clear, specific, and immediately actionable response; lower scores reflect responses that are incomplete, unclear, or fail to move the user forward, down to 0 for a response that provides no useful guidance at all.

The Factual Consistency, Correctness, and Helpfulness scores are produced by a panel of three large language models acting as judges (GPT-4o, GPT-4.1-Mini, GPT-4.1); the panel spans different capacities within the GPT family to verify that scores are robust to judge capability. All three judges receive the same evaluation inputs, including the Knowledge Base supplied to the agent at session time. The reported score is the mean across the three judges; the full prompt, per-rubric inter-judge agreement, and a discussion of single-provider limitations are in Appendix \ref{appendix:llm_as_judge} and §\ref{sec:discussion}.

\subsubsection{Question Bank} 
To collect realistic user queries for the ablation evaluation, we conducted a Wizard of Oz study with four participants who were not part of the user study reported in §6. During each session, the participant assembled a LACK Coffee Table, one of the three furniture types in the evaluation set (§5.1.2). Responses to participant queries were delivered by a human operator from a prescripted answer set rather than by the live ProcAgent. Decoupling query collection from the live system ensures that the queries reflect task-driven user confusion rather than ProcAgent's specific response behavior, making the question bank a stable evaluation set across configurations.

We captured 53 raw utterances across the four sessions. Each utterance corresponds to a single query addressed to the operator, segmented at speaker turns. Unlike traditional benchmarks that filter for uniqueness, we preserved repeated and closely paraphrased queries (e.g., what do I do next? uttered at multiple points in a single session) because their recurrence at different assembly milestones is informative about how user confusion is distributed across the task. The 53 utterances reduce to 20 unique query templates, where two utterances share a template if they differ only in surface phrasing or referenced object.

The 20 templates were grouped into three categories. The categories emerged from inductive coding of the utterances rather than from a target set defined in advance. The three categories are Procedural Guidance, State Validation, and Progress and History; counts and example queries are in Table~\ref{tab:query_distribution}.
We hypothesize that the three categories map to the three response paths ProcAgent uses to answer queries: Procedural Guidance queries to the task graph alone, State Validation queries to the visual verifier, and Progress and History queries to the Session Context and Knowledge Base. The Routing F1 metric (§5.1.4) tests this mapping by evaluating whether the procedural agent correctly routes each query template to its appropriate response path.


\begin{table}[ht]
\centering
\caption{%
User query categories from the Wizard of Oz study (53 utterances across 4 participants, reducing to 20 unique templates). The categories emerged inductively from the collected utterances.
}
\label{tab:query_distribution}
\resizebox{\textwidth}{!}{%
\begin{tabular}{lccl}
\toprule[2pt]
\textbf{Category} & \textbf{Unique} & \textbf{Total Freq.} & \textbf{Contextual Examples \& Responses} \\ \midrule
\textbf{Procedural Guidance} & 7 & 22 & \cellcolor[gray]{0.9} \textbf{Q1:} What is the next step? \\
\textit{(Future actions / FSM)} & & & \cellcolor[gray]{0.9} \textbf{A1:} Align the leg with the corner bracket. \\ 

& & & \cellcolor[gray]{0.9} \textbf{Q2:} Where should I put this piece? \\
& & & \cellcolor[gray]{0.9} \textbf{A2:} Insert the shelf into the side grooves. \\ \midrule

\textbf{State Validation} & 9 & 16 & \cellcolor[gray]{0.9} \textbf{Q1:} Am I doing this correctly? \\
\textit{(Current state / Verifier)} & & & \cellcolor[gray]{0.9} \textbf{A1:} Yes, the alignment looks perfect. \\ 

& & & \cellcolor[gray]{0.9} \textbf{Q2:} Is the alignment correct? \\
& & & \cellcolor[gray]{0.9} \textbf{A2:} No, rotate the leg 90 degrees. \\ \midrule

\textbf{Progress \& History} & 4 & 15 & \cellcolor[gray]{0.9} \textbf{Q1:} How many steps are left? \\
\textit{(Context / RAG)} & & & \cellcolor[gray]{0.9} \textbf{A1:} You have one last step. Attach the shelf that you just picked up. \\

& & & \cellcolor[gray]{0.9} \textbf{Q2:} Give me a recap of my work. \\
& & & \cellcolor[gray]{0.9} \textbf{A2:} You have finished attachment of two legs. Two more to go. \\ \bottomrule[2pt]
\end{tabular}%
}
\end{table}

\subsection{End-to-End System Comparison}
To contextualize the benefits of \sys, we compare it against a \textit{Single VLM} (llama-3.1-8b  at $Q5\_K\_M$ quantization) baseline that removes the agentic structure and relies on a single vision-language model invocation. Given the user's query and the five most recent camera frames, this baseline produces a response directly, without explicit state tracking, retrieval, visual verification, or human feedback. This provides a simple but informative point of comparison for assessing what the full architecture contributes.

As shown in Table \ref{tab:one_vlm}, \sys outperforms the Single VLM baseline across all response-quality measures. Factual consistency improves from $2.54 \pm 1.51$ to $3.71 \pm 1.18$, helpfulness from $1.56 \pm 1.30$ to $3.43 \pm 1.38$, and correctness from $1.85 \pm 1.40$ to $3.64 \pm 1.35$. These differences suggest that assembly guidance benefits substantially from explicit procedural modeling. Without access to verified state history, task constraints, or structured retrieval, the Single VLM can only interpret a short visual window and the current query. As a result, its responses are often locally plausible but insufficiently grounded in the user’s actual progress through the task.

The Single VLM baseline is faster on average ($7.35 \pm 0.24$\,s versus $8.02 \pm 0.67$\,s for \sys), which is expected given that it avoids the additional coordination and verification steps of the full system. However, this latency advantage is modest relative to the gains in response quality. Taken together, these results indicate that the additional structure in \sys introduces limited overhead while substantially improving factual consistency, correctness, and practical usefulness.

\begin{table}[ht]
\centering
\caption{End-to-end comparison between \sys and a Single VLM baseline. The Single VLM baseline responds directly from the user query and the five most recent camera frames, without explicit state tracking or other agentic components. \sys substantially improves factual consistency, helpfulness, and correctness, with only a modest increase in response time. Values are mean $\pm$ standard deviation.}
\label{tab:one_vlm}
\definecolor{best}{gray}{0.9}
\resizebox{\textwidth}{!}{%
\begin{tabular}{lccccccc}
\toprule[2pt]
\textbf{System}  
  & \textbf{Factual Consistency\,$\uparrow$} 
  & \textbf{Helpfulness\,$\uparrow$}
  & \textbf{Correctness\,$\uparrow$}
  & \textbf{Avg. Response (s)\,$\downarrow$} \\
\midrule
Single VLM       & 2.54\,$\pm$\,1.51 & 1.56\,$\pm$\,1.30 & 1.85\,$\pm$\,1.40 & \cellcolor{best}\textbf{7.35\,$\pm$\,0.24} \\

\midrule
\textbf{\sys}   
  & \cellcolor{best}\textbf{3.71\,$\pm$\,1.18}  
  & \cellcolor{best}\textbf{3.43\,$\pm$\,1.38}
  & \cellcolor{best}\textbf{3.64\,$\pm$\,1.35}
  & 8.02\,$\pm$\,0.67          \\
\bottomrule[2pt]
\end{tabular}%
}
\end{table}

\subsection{Component Selection}
\label{sec:component_sel}
In this section, we evaluate and finalize the configuration of two components: the proposer (backbone model and temporal smoothing parameters) and the procedural agent (language model). We select each through a comparison against alternatives.

\subsubsection{Proposer Backbone Selection}

\begin{table}[H]
    \centering
    \caption{Performance comparison of various action recognition methods for furniture assembly guidance. We report the frame-level accuracy (Top-1 and Top-3) and Macro-F1 score for each configuration. Note that CLIP-based variants are tested in few-shot settings.}
    \label{tab:clip_vs_others_results} 
    \resizebox{0.35\textwidth}{!}{
    \definecolor{best}{gray}{0.9} 
    \begin{tabular}{l c c c c}
        \toprule[2pt]
        \textbf{Method} & \textbf{Shots} & \multicolumn{2}{c}{\textbf{Frame Acc}} & {\textbf{Macro-F1}} \\
        \cmidrule(lr){3-4}
        & & {Top 1} & {Top 3} & \\
        \midrule
        ResNet18 & -- & 33.19 & 58.61 & 17.21 \\
        ResNet34 & -- & 35.76 & 63.29 & 18.64 \\
        ResNet50 & -- & 39.53 & 66.34 & 20.37 \\
        C3D & -- & 43.97 & 72.16 & 31.27 \\
        P3D & -- & 51.24 & 82.94 & 39.41 \\
        \midrule
        \multirow{4}{*}{CLIP} & 8 & 35.12 & 53.24 & 17.53 \\
        & 16 & 43.64 & 64.61 & 32.41 \\
        & 32 & 57.46 & 81.46 & 40.59 \\
        & 64 & \cellcolor{best}\textbf{65.24} & \cellcolor{best}\textbf{89.26} & \cellcolor{best}\textbf{43.31} \\
        \bottomrule[2pt]
    \end{tabular}
    }
\end{table}

We compare CLIP against a range of supervised action-recognition models on frame-level classification of assembly steps. Frame-level Top-1, Top-3, and Macro-F1 are reported for each method in Table~\ref{tab:clip_vs_others_results}. CLIP variants are evaluated in a few-shot setting; the supervised baselines are trained on the full task-specific training set. Among the supervised baselines, performance scales with model capacity. ResNet variants reach 33.19\% to 39.53\% Top-1 accuracy. Temporal models perform more competitively: C3D reaches 43.97\% and P3D reaches 51.24\% Top-1, benefiting from their ability to capture short-range temporal dynamics. CLIP at 64 shots surpasses all supervised baselines across every metric, achieving 65.24\% Top-1, 89.26\% Top-3, and 43.31 Macro-F1, despite operating without any task-specific supervised training. Beyond raw accuracy, CLIP's few-shot adaptability (text descriptions plus a small number of visual examples are sufficient to accommodate new procedural steps) and its low per-frame inference latency make it well suited for continuous frame-level monitoring on the Jetson AGX Orin. We adopt CLIP at 64 shots as the proposer backbone for all subsequent experiments.

\subsubsection{Perception-Stack Smoothing and Frame Selection}

With CLIP fixed as the backbone, we search over the proposer's temporal smoothing parameters and the visual verifier's frame-selection strategy. We vary three parameters: the sliding-window size $W$, which controls how many consecutive frames must agree on a state before a candidate is emitted; the stride $S$, which controls the window's step size between emissions; and a check-timeout threshold (None, 15s, 20s, 25s) that triggers a user-confirmation event when no state change has been detected within the timeout, to confirm that the user is genuinely still in the same state. We additionally compare three frame-selection strategies for the verifier: Single Frame; Multi Frame (0--100\%), which samples the first, middle, and last frames of the window; and Multi Frame (50--100\% RWS), which samples the middle, third-quarter, and last frames. State Accuracy and Sequence Adherence for the full search are reported in Appendix~B (Table~\ref{tab:hyper_parameter_ablation}).

We observe that window sizes that are too large reduce adherence by slowing candidate emission, making the system less responsive to genuine state changes. Window sizes that are too small introduce instability: candidates are emitted too readily, leading to noisy transitions and degraded accuracy. The configuration
$W=30$, $S=15$, $Check=20s$ strikes the best overall balance, achieving State Accuracy of $78.25 \pm 9.49\%$ under the single frame condition and sequence adherence of $74.28 \pm 27.48\%$ under Multi Frame (50--100\% RWS), among the highest adherence scores in the search. While $W=10$, $S=5$, $Check=25$s achieves a marginally higher peak accuracy, we select 
$W=30$, $S=15$, $Check=20$s with the Multi Frame (50--100\% RWS) verification strategy for its superior stability and more consistent multi-frame grounding. This configuration is fixed for all subsequent experiments.

\begin{table}[ht]
\centering
\caption{%
  Reasoning Agent Ablation: Impact of Central LLM Choice on System Performance.
  All configurations use the full \sys~pipeline (CLIP + Verifier + FSM + RAG).
  Models are evaluated in quantized form on the Jetson AGX Orin.
}
\label{tab:ablation_agent}
\definecolor{best}{gray}{0.9}
\resizebox{\textwidth}{!}{%
\begin{tabular}{lccccccc}
\toprule[2pt]
\textbf{Agent Model}  
  & \textbf{Params}  
  & \textbf{Tool Use F1\,$\uparrow$}
  & \textbf{Factual Consistency \,$\uparrow$} 
  & \textbf{Helpfulness\,$\uparrow$}
  & \textbf{Correctness\,$\uparrow$}
  & \textbf{Avg. Response (s)\,$\downarrow$} \\
\midrule
Phi-3-Mini (Microsoft)        & 3.8B  & 0.2642 & 3.16\,$\pm$\,1.12 & 2.81\,$\pm$\,1.21 & 2.92\,$\pm$\,1.23 & \cellcolor{best}\textbf{2.16\,$\pm$\,0.34} \\

LLaMA-3.1-Instruct (Meta)        & 8B  & 0.5240 & 3.47\,$\pm$\,1.19 & 3.15\,$\pm$\,1.39 & 3.33\,$\pm$\,1.38 & 6.1\,$\pm$\,0.28 \\

\midrule
\textbf{GPT-OSS-20B (Ours)}   
  & \textbf{20B}   
  & \cellcolor{best}\textbf{0.5664} 
  & \cellcolor{best}\textbf{3.71\,$\pm$\,1.18}  
  & \cellcolor{best}\textbf{3.43\,$\pm$\,1.38}
  & \cellcolor{best}\textbf{3.64\,$\pm$\,1.35}
  & 8.02\,$\pm$\,0.67          \\
\bottomrule[2pt]
\end{tabular}%
}
\end{table}

\subsubsection{Agent Backbone}
Table~\ref{tab:ablation_agent} reports results for the three agent backbones described in §5.1.3 (Phi-3-Mini-4k-Instruct, LLaMA-3.1-Instruct, and GPT-OSS-20B), each evaluated under the full ProcAgent configuration. GPT-OSS-20B leads on every quality dimension, achieving the highest Factual Consistency ($3.71 \pm 1.18$), Helpfulness ($3.43 \pm 1.38$), Correctness ($3.64 \pm 1.35$), and Routing F1 (0.57). Phi-3-Mini delivers the fastest response time at $2.16 \pm 0.34$
 seconds, but its Routing F1 of $0.26$ indicates that it frequently fails to invoke the visual verifier at the right moment, which undermines the agent's coordination of the rest of the pipeline. LLaMA-3.1-Instruct sits between the two on quality metrics and slightly above GPT-OSS-20B on response time. The pattern is consistent across metrics: agent capacity correlates with response quality and routing accuracy, while smaller agents trade quality for latency.

\begin{table*}[ht]
\centering
\caption{
  Architectural Ablation Study: Contribution of Each System Component. 
  All configurations include \textbf{Procedural Agent and Proposer}; we vary remaining components. Results are averaged over 15 assembly sessions (IKEA LACK Coffee Table, LACK Side Table, Lack TV Bench).
  $\uparrow$ = higher is better, $\downarrow$ = lower is better.
  $\pm$ denotes one standard deviation across sessions. \textbf{The gray shaded area represents the best result and the underlines represent the second best result for each metric.}
}
\label{tab:ablation_results}
\definecolor{best}{gray}{0.9} 
\resizebox{\linewidth}{!}{
\begin{tabular}{lccccc|ccccccc}
\toprule[2pt]
& \multicolumn{5}{c}{\textbf{Configurations}} & \multicolumn{7}{c}{\textbf{Performance Metrics}} \\
\cmidrule(lr){2-6} \cmidrule(lr){7-13}
\textbf{Version} & \textbf{\begin{tabular}[c]{@{}c@{}}Verifier \\(State)\end{tabular}} & \textbf{\begin{tabular}[c]{@{}c@{}}Finite-State\\ Task Model\end{tabular}} & \textbf{\begin{tabular}[c]{@{}c@{}}Knowledge\\ Base\end{tabular}} & \textbf{\begin{tabular}[c]{@{}c@{}}Verifier \\(Query)\end{tabular}} & \textbf{\begin{tabular}[c]{@{}c@{}}User \\Confirmation\end{tabular}} & \textbf{Acc. $\uparrow$} & \textbf{Adh. $\uparrow$} & \textbf{Fact. $\uparrow$} & \textbf{help. $\uparrow$} & \textbf{correct. $\uparrow$} & \textbf{Resp. (s) $\downarrow$} & \textbf{Pred. (s) $\downarrow$} \\ 
\midrule
A & \xmark & \xmark & \xmark & \xmark & \xmark & \cellcolor{best}\textbf{87.9\,$\pm$\,9.3} & 42.9\,$\pm$\,26.7 & 3.39\,$\pm$\,1.43 & 2.72\,$\pm$\,1.39 & 3.06\,$\pm$\,1.53 & \cellcolor{best}\textbf{2.49\,$\pm$\,0.38} & \cellcolor{best}\textbf{0.91\,$\pm$\,0.03} \\
B & \cmark & \xmark & \xmark & \xmark & \xmark & 82.6\,$\pm$\,10.3 & \underline{47.1\,$\pm$\,30.0} & 3.54\,$\pm$\,1.38 & 2.88\,$\pm$\,1.34 & 3.26\,$\pm$\,1.50 & 2.66\,$\pm$\,0.30 & 3.60\,$\pm$\,0.03 \\
C & \cmark & \cmark & \xmark & \xmark & \xmark & \underline{86.0\,$\pm$\,8.2} & 39.4\,$\pm$\,32.5 & 3.46\,$\pm$\,1.43 & 2.82\,$\pm$\,1.37 & 3.22\,$\pm$\,1.55 & 2.71\,$\pm$\,0.34 & 1.89\,$\pm$\,1.25 \\
D & \cmark & \xmark & \cmark & \xmark & \xmark & 83.1\,$\pm$\,10.3 & 46.5\,$\pm$\,30.5 & \underline{3.89\,$\pm$\,1.33} & 3.18\,$\pm$\,1.44 & 3.57\,$\pm$\,1.52 & 2.70\,$\pm$\,0.63 & 2.77\,$\pm$\,1.21 \\
E & \cmark & \cmark & \cmark & \xmark & \xmark & 85.6\,$\pm$\,7.6 & 40.4\,$\pm$\,34.1 & \cellcolor{best}\textbf{3.92\,$\pm$\,1.35} & 3.23\,$\pm$\,1.44 & 3.63\,$\pm$\,1.52 & \underline{2.53\,$\pm$\,0.61} & 1.89\,$\pm$\,1.25 \\
F & \cmark & \cmark & \cmark & \cmark & \xmark & 85.5\,$\pm$\,7.9 & 39.4\,$\pm$\,32.5 & 3.8\,$\pm$\,1.13 & \cellcolor{best}\textbf{3.44\,$\pm$\,1.32} & \cellcolor{best}\textbf{3.69\,$\pm$\,1.30} & 6.15\,$\pm$\,3.04 & \underline{1.06\,$\pm$\,0.08} \\
\textbf{\sys} & \cmark & \cmark & \cmark & \cmark & \cmark & 83.7\,$\pm$\,8.5 & \cellcolor{best}\textbf{72.8\,$\pm$\,18.9} & 3.71\,$\pm$\,1.18 & \underline{3.43\,$\pm$\,1.38} & \underline{3.64\,$\pm$\,1.35} & 8.02\,$\pm$\,0.67 & 1.10\,$\pm$\,0.09 \\
\bottomrule[2pt]
\end{tabular}
}
\end{table*}

 \subsection{System Ablation}
 \label{sec:ablation}
 
The system ablation evaluates each component's marginal contribution to procedural control and response quality. We run the seven configurations described in §5.1.3 across the 15 evaluation sessions. Because the IKEA demonstration dataset does not include user-induced sequencing errors, user confirmation is simulated using the rule-based mechanism described in §5.1.3, which activates only when the system makes a sequencing error. Results are reported in Table~\ref{tab:ablation_results}.

\textbf{Effect of the Visual Verifier on State Tracking.} 
 Configuration A (the proposer alone) achieves the highest raw State Accuracy at $87.9 \pm 9.3\%$, but one of the lowest Sequence Adherence scores at $42.9 \pm 26.7\%$. The proposer often emits visually plausible candidates without enforcing procedural order. Adding the visual verifier in Configuration B improves Sequence Adherence to $47.1 \pm 30.0\%$ and also lifts Factual Consistency and Correctness, indicating that visual verification contributes beyond classification accuracy by stabilizing the system's transition decisions.

\textbf{Effect of the Finite-State Task Model.}
 Adding the task model in Configuration C reduces Sequence Adherence to $39.4 \pm 32.5\%$. This may appear counterintuitive but is consistent with how the task model operates without a confirmation mechanism: the task model enforces guard conditions strictly, and when the visual signal is ambiguous, it rejects valid transitions rather than committing to uncertain state updates. The task models's contribution is therefore not visible in Sequence Adherence alone; it is visible in the task model's role as a gating mechanism for verifier invocation, which produces measurable reductions in GPU load, power consumption, and thermal stress. We report these effects in §\ref{sec:resource_thermal}.

\textbf{Effect of the Knowledge Base.} 
 Configurations D and E (with Knowledge Base retrieval) improve response quality. They achieve the highest Factual Consistency scores in the ablation: $3.89 \pm 1.33$ and $3.92 \pm 1.35$ respectively. Correctness also improves to $3.57 \pm 1.52$ in D and $3.63 \pm 1.52$ in E. Without retrieval, the agent recognizes the current state but lacks step-specific procedural detail; with retrieval, responses are anchored in the requirements and completion criteria of the current step.

\textbf{Effect of the Visual Verifier on Query Grounding.} 
 Configuration F adds the verifier on the user-query path. It produces the strongest Helpfulness ($3.44 \pm 1.32$) and Correctness ($3.69 \pm 1.30$) scores in the ablation. Spatially contingent queries about part orientation, alignment, or completeness are difficult to answer from textual context and state information alone; visual grounding makes these responses concrete. Sequence Adherence in F remains at $39.4 \pm 32.5\%$, consistent with the pattern that perception and retrieval components do not by themselves enforce procedural order.

 \textbf{Effect of User Confirmation.} 
 The full \sys configuration adds user confirmation to Configuration F. Sequence Adherence rises from 39.4\% in F to $72.8 \pm 18.9\%$ in \sys, a 33-point jump that is the largest single contribution in the ablation. This identifies user confirmation as the mechanism that resolves the architectural disagreement between the visual verifier and the task model: when the verifier confirms a candidate but the task model's guard rejects it, neither the perception evidence nor the procedural structure can adjudicate alone, and user confirmation supplies the missing signal. The improvement comes with a modest cost: average Response Time rises to $8.02 \pm 0.67$ seconds, due to the additional confirmation turn.

 Taken together, the ablation reveals an asymmetric contribution structure across metrics. Each component improves a different aspect of system behavior: the visual verifier improves response correctness; the Knowledge Base improves factual grounding; the task model provides procedural structure and resource efficiency (§5.4); the visual verifier on the query path produces the strongest spatial grounding for user questions; and user confirmation drives Sequence Adherence by resolving disagreements between the verifier and the task model. No single configuration improves every metric monotonically, and the full \sys achieves the strongest combined performance because each component is responsible for a different failure mode.

 \begin{figure}[!htb]
    \centering
    \includegraphics[trim= 10 145 0 100, clip, width=\textwidth]{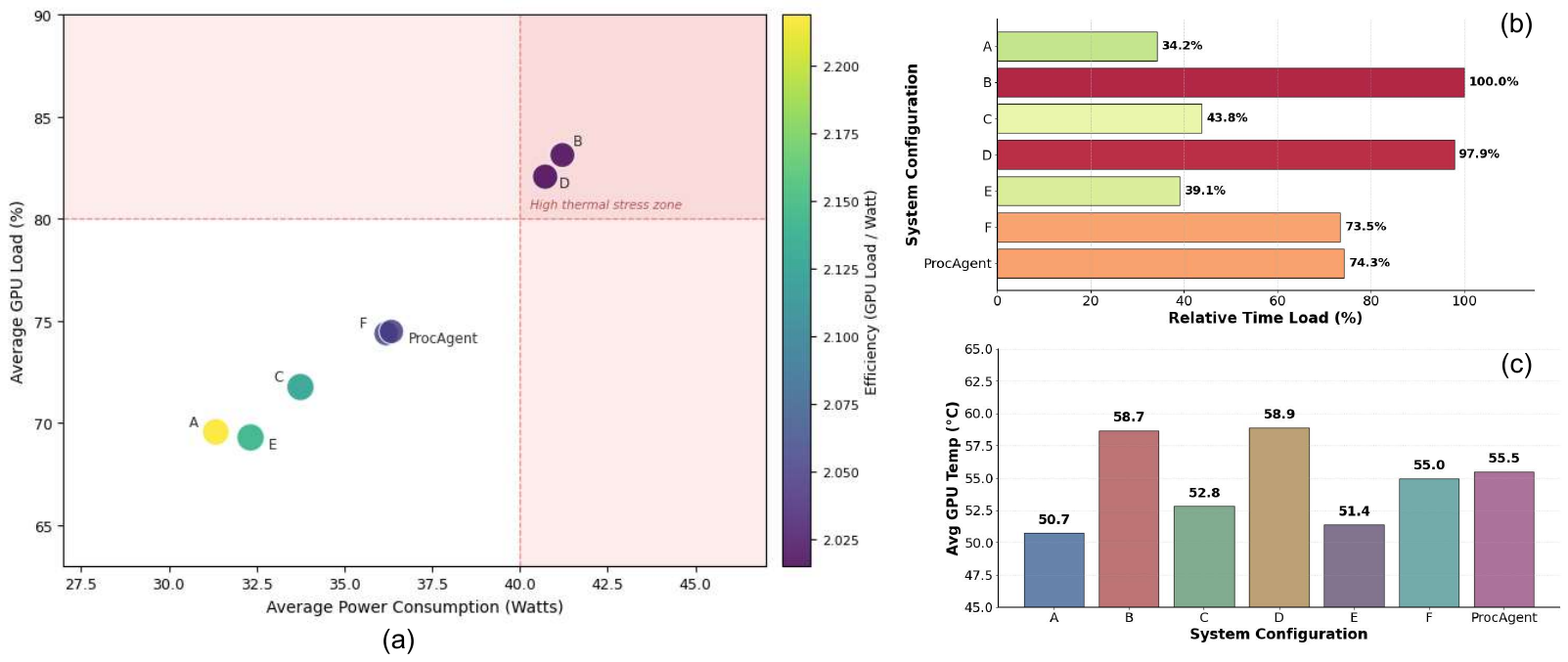}
    \caption{\textbf{Hardware Performance and Thermal Analysis}. Comparative benchmarking of ablation configurations (A–F) and the integrated system (\sys) on the NVIDIA Jetson AGX Orin platform. \textbf{(A)} Power-Efficiency Trade-offs: Scatter plot showing the relationship between GPU load and power consumption. The dashed red boundaries define a high thermal stress region (Power > 40W, Load > 80\%), where configurations C and E reside. The "Full Sys" configuration achieves a balanced operating point, optimizing throughput without triggering thermal safety limits. \textbf{(B)} Relative Computational Time Load: Horizontal bar chart displaying the normalized duration of processing sessions relative to the maximum recorded baseline (Configuration C). Configurations G and Full Sys demonstrate consistent temporal footprints (~74\%) required for robust multimodal reasoning during procedural assembly. \textbf{(C)} Thermal Profile Distribution: Mean GPU temperatures recorded across all experimental trials. High-performance configurations (C, E) approach the 60°C threshold, highlighting the necessity of the efficient resource management strategies employed in the finalized system (55.5°C) to ensure long-term hardware reliability.}
    \label{fig:edge_metrics}
    \Description[]{}
    \vspace{-2mm}
\end{figure}

 \subsection{Resource and Thermal Profile}
 \label{sec:resource_thermal}
We characterize the on-device resource cost of each configuration to evaluate whether the architectural choices identified in §\ref{sec:component_sel} translate into measurable runtime savings on the Jetson AGX Orin. Three measurements are reported in Figure~\ref{fig:edge_metrics}: average GPU load against power consumption (Figure~\ref{fig:edge_metrics}a), session duration normalized to the heaviest configuration (Figure~\ref{fig:edge_metrics}b), and mean GPU temperature (Figure~\ref{fig:edge_metrics}c).

\textbf{Power and GPU Load.} 
Figure~\ref{fig:edge_metrics}a shows that Configurations B and D (visual verifier without task model) sit inside the high-thermal-stress region, defined as average GPU load above 80\% combined with average power above 40W. In these configurations, every candidate emitted by the proposer is forwarded to the visual verifier without filtering. Configurations that include the FSM (C, E, F, and \sys) sit outside this region. The task model reduces verifier calls because candidates that do not match a guard-satisfiable event from the current task state are filtered before verification, removing the GPU load associated with those calls. The full \sys operates at roughly 75\% GPU load and 41W average power, well within the safe operating envelope of the Jetson platform.

 \textbf{Session Duration.} 
 Figure~\ref{fig:edge_metrics}b reports session duration as a percentage of Configuration B's duration, the heaviest baseline. Configurations with the task model are substantially shorter: Configuration C runs at 43.8\% of the baseline, Configuration E at 39.1\%, and the full \sys at 74.3\%. The \sys footprint is higher than the task model-only configurations because user confirmation adds a confirmation-dialogue turn to a subset of sessions, but it remains far below the task model-excluded baselines. Configurations without the task model (B and D) take roughly twice as long as their task model-included counterparts, because each candidate requires a verifier call rather than being filtered upstream.

 \textbf{Thermal Profile.} 
 Figure~\ref{fig:edge_metrics}c reports mean GPU temperature across all experimental trials. Configurations B and D approach 59$^\circ$C, the highest temperatures in the study and within a few degrees of the thermal threshold beyond which the Jetson begins dynamic frequency scaling. Configurations with the task model operate between 51$^\circ$C and 55$^\circ$C, and the full \sys runs at 55.5$^\circ$C. Sustained operation in the high-thermal regime would, over long sessions, trigger throttling and reduce inference throughput; the task model-included configurations avoid this regime by construction.

The resource profile supports the architectural argument made in §5.4: the task model is not only a procedural-control component but also a runtime efficiency mechanism. Filtering candidates against guard conditions reduces verifier invocations, which in turn reduces GPU load, power draw, session duration, and thermal stress. The full \sys achieves the system's procedural and quality wins while remaining within the Jetson's safe operating envelope.

\section{User Studies}
To complement the quantitative ablation study, we conducted a user study to evaluate ProcAgent from the perspective of the people it is designed to help. While the ablation study measures system-level performance: state accuracy, sequence adherence, hallucination rate, the user study asks a different and equally important question: does the system actually help real users assemble furniture more successfully, and do they find it usable and trustworthy? This section describes the study design, tasks, participants, metrics, procedure, and results.

\subsection{Design}

The study followed a within-subjects design in which each participant completed an Lack Coffee Table assembly task with \sys~. Each session was conducted in a controlled indoor environment using the same hardware setup described in Section \ref{ssubsec:hardware}: a Jetson AGX Orin running the full system pipeline, a Logitech C920s Pro camera, an EMEET Conference Speaker, and a directional microphone. Participants were not shown any technical details of the system and were simply told they would be assembling furniture with the help of an AI assistant.

\subsection{Task}

Participants were asked to assemble the Lack Coffee table IKEA furniture from scratch using only the physical components provided; no instruction manual was given. The assembly task was selected from the same furniture categories used in the ablation study, covering a representative range of sub-assembly complexity and step count. The same task was used across all participants to ensure comparability of results.

\subsection{Participant}
We recruited 10 participants (2 female, 8 male) through convenience sampling within our institution.  No participant had prior experience with AI-assisted assembly tools. All participants provided informed consent before the study began and were free to withdraw at any time.

\subsection{Metric}

\textbf{Post-Session User Experience Survey.} After completing the session, each participant filled out a custom 8-item post-session survey evaluating their subjective experience with \sys. Each item was rated on a 5-point Likert scale ranging from Strongly Disagree to Strongly Agree, covering eight dimensions of the interaction experience:

\begin{itemize}
    \item (Q1) Comprehensibility: Whether the agent's feedback was clear and easy to understand.
    \item (Q2) Actionability: Whether the instructions provided concrete, immediately actionable next steps.
    \item (Q3) Terminology: Whether the agent's wording aligned naturally with furniture assembly language.
    \item (Q4) Confidence: Whether the agent's feedback increased the participant's confidence during assembly.
    \item (Q5) Mental Demand: Whether processing the agent's feedback felt mentally demanding.
    \item (Q6) Responsiveness: Whether feedback was timely and responsive.
    \item (Q7) Privacy: Whether knowing that all processing was done locally on-device made participants feel comfortable.
    \item (Q8) Trust: Whether participants trusted the agent's proactive "Hold on" interventions.
\end{itemize}

    

\subsection{Procedure}
Each study session followed a structured protocol and lasted approximately 15 minutes in total. Upon arrival, participants were given a brief overview of the study  and a two-minutes familiarization period during which the system's interaction model was briefly explained, specifically, that they could ask the agent questions at any time and that the agent might occasionally speak to them unprompted if it detected something worth flagging. No further instructions were given. They were then introduced to the physical assembly components but given no instructions on how to proceed.

After completing each session, participants filled out the user experience evaluation form. Sessions were video recorded for post-hoc analysis, and all recordings were stored locally on the study device in accordance with our institution's data handling guidelines. The study was approved by the authors’ institutional IRB (Approval number: IRB-25-09024-XP), and all participants provided informed consent.

\subsection{Results}

\begin{figure}[ht]
    \centering
    \includegraphics[trim= 10 20 30 20, clip, width=\textwidth]{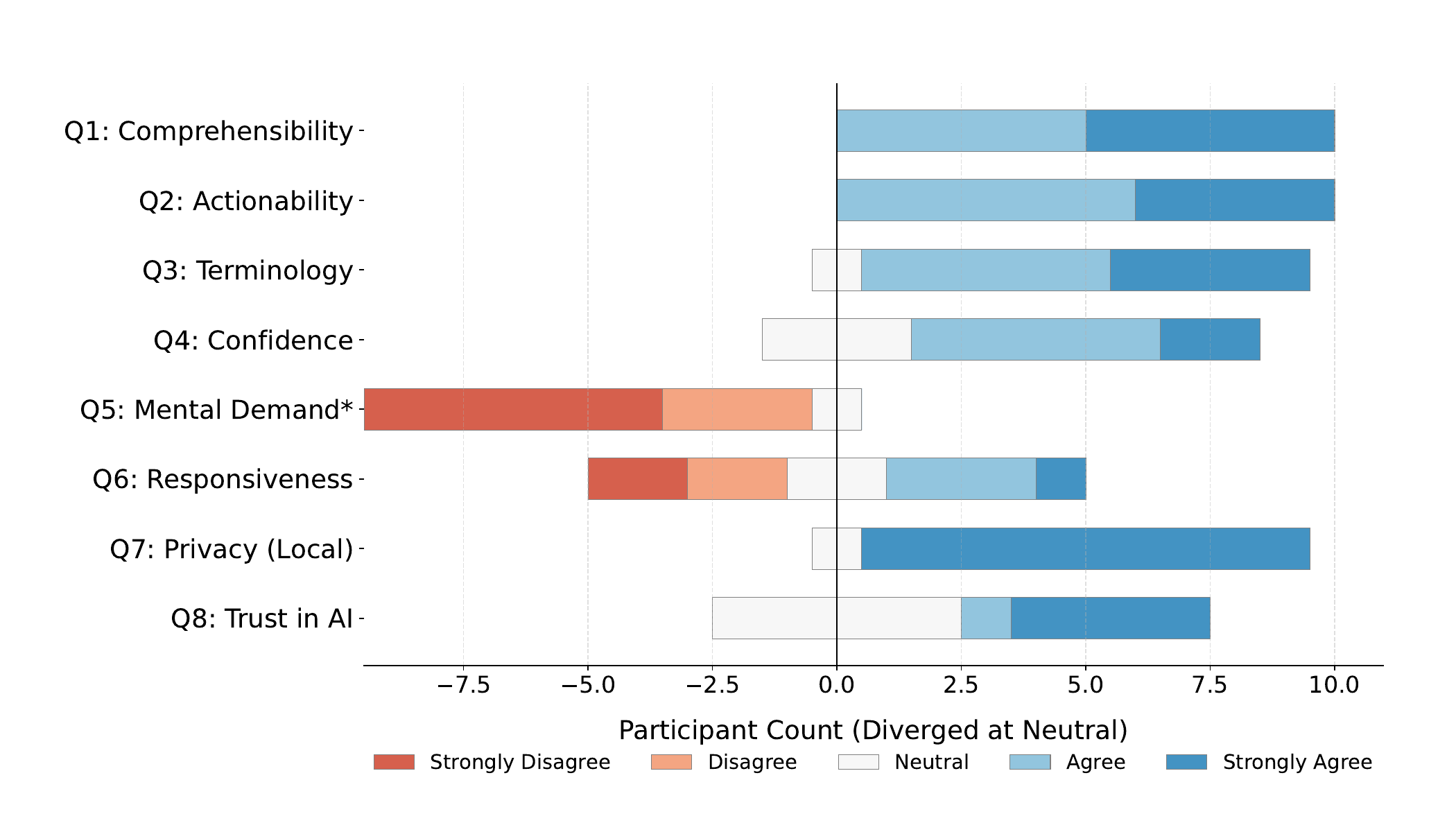}
    \caption{\textbf{Likert-scale user study results for ProcAgent (n = 10)}. After using the system, participants completed an 8-item, 5-point questionnaire. The diverging stacked bar chart shows generally positive ratings across most usability and perception dimensions, with especially strong responses for comprehensibility, actionability, confidence, and privacy.}
    \label{fig:user_exp}
    \Description[]{}
    \vspace{-2mm}
\end{figure}

Post-Session User Experience Survey. The survey results, summarized in Figure \ref{fig:user_exp}, reveal a consistently positive reception across most dimensions. Comprehensibility (Q1) and Actionability (Q2) received the strongest endorsement, with all participants rating the agent's feedback as clear and actionable, reflecting the effectiveness of the agent's context-aware, FSM-grounded response generation in producing guidance that is specific and immediately usable rather than generic.

Terminology alignment (Q3) and Privacy comfort (Q7) also received strong positive ratings, with the large majority of participants agreeing or strongly agreeing. The strong Privacy result is particularly noteworthy: participants responded positively to knowing that all inference occurred locally on the device, suggesting that on-device deployment is not merely a technical design choice but a meaningful contributor to user trust and comfort.

Confidence (Q4) and Trust in interventions (Q8) showed broadly positive results, though with slightly more neutral responses than Q1 and Q2. This pattern is consistent with what one might expect from the first-time users of an AI assembly assistant. Participants were generally willing to follow the agent's guidance but retained some residual uncertainty, particularly around the proactive interventions, which represent a novel and unfamiliar interaction pattern for most users.

Mental Demand (Q5) was phrased as a negative item, meaning that disagreement reflects a positive outcome. The strong skew toward Strongly Disagree indicates that the majority of participants did not find the system mentally taxing during assembly. This validates two key design decisions: the agent's deliberate restraint in speaking only when genuinely needed, and the Progressive Disclosure strategy of delivering a lightweight acknowledgment, accompanied by a brief waiting sound, before the full response. Together, these kept cognitive load comfortably low even during the most demanding moments of the assembly task.

Responsiveness and Feedback Time (Q6) received a broadly positive response, with most participants finding the system's response timing acceptable. The small number of neutral responses is understandable given that visually grounded responses can take 8-12 seconds to generate. The combination of an immediate verbal acknowledgment and a waiting sound appears to have effectively bridged this gap for most users, making the delay feel like a natural deliberative pause rather than a system lag. That said, the neutral responses suggest that response latency remains perceptible to some users and is a worthwhile target for future optimization.

\section{Discussion and Limitations}
\label{sec:discussion}

Our findings suggest that \sys is best understood as a tightly coupled interactive system rather than a set of independent modules. In the ablation, adding individual components sometimes reduced performance before the full pipeline was in place, indicating that incremental ablations can understate the value of interdependent mechanisms such as FSM tracking, visual verification, and human feedback.

More broadly, the results show that human feedback serves as a primary source of state disambiguation rather than a fallback when perception fails. User confirmations anchor the tracker to reliable ground truth and support more confident state transitions. The gains in sequence adherence therefore reflect not only improved perception, but also the benefit of explicitly integrating user input into procedural tracking.

The system also shows how edge deployment constraints can productively shape design. Running on a single Jetson AGX Orin motivated staged verification, FSM-based gating, and Progressive Disclosure, which together limit unnecessary model invocations and help manage latency. At the same time, our results underscore that perception accuracy and procedural correctness are distinct: a system may recognize the current state correctly while still failing to guide the user through the correct next action. In such settings, explicit task structure is as important as visual recognition. The user study further suggests that trust depends not only on correctness, but also on selective intervention behavior.

These findings should be interpreted in light of several limitations. We evaluated \sys on 15 demonstration videos across three IKEA furniture categories; broader evaluation on assemblies with longer horizons, greater branching, or more visually similar parts would better establish generality. Although we use three LLM judges of varying capability, all three -- GPT-4o, GPT-4.1-Mini, and GPT-4.1 -- are from the same provider and share underlying training data, alignment procedures, and potential biases. Scores that appear robust across the panel may still reflect systematic OpenAI-family tendencies rather than true inter-model independence. Future evaluations should incorporate judges from distinct model families, such as Claude or Gemini, to provide stronger evidence of score robustness across genuinely diverse evaluators. In the ablation, human feedback was simulated with a logic-based trigger to isolate its functional role, but larger studies with real users would better capture variation in actual confirmations and mistakes. The current FSTM supports flexible ordering among interchangeable sub-steps but assumes forward progress and does not support undo or backward revision of previously confirmed steps. Extending \sys to a new category also requires an offline authoring pipeline for step extraction, validation, and hierarchical context generation, which is amortized across sessions but still affects scalability. Finally, the current implementation relies on a single fixed camera and assumes one active sub-step at a time, leaving occlusion and concurrent sub-step execution for future work.

\section{Conclusion}

We presented \sys, a fully on-device agentic framework for real-time procedural assembly guidance, combining a conversational agent, a two-stage perception stack, an FSM-grounded task model, a RAG-based knowledge base, and a Human-in-the-Loop confirmation mechanism -- all running locally on a single NVIDIA Jetson AGX Orin.
The central finding is that reliable procedural guidance requires more than accurate perception. It requires knowing what is legally possible at each step, when visual verification is genuinely needed, and when to defer to the human. No partial configuration achieves all three simultaneously -- only the full system does, and the 33-point adherence gain over the baseline reflects that. Each component covers a distinct lackings of others: the Visual Verifier catches what the Low-Latency Proposer misses, the FSM enforces what the Verifier cannot, RAG grounds what the FSM does not know, the query grounded visual context confirms what text alone cannot deliver, and Human Feedback resolves what no model can determine alone.

User study participants responded positively across nearly all dimensions. Feedback was rated as clear and immediately actionable by all participants, and the system's use of assembly-specific terminology felt natural. Privacy comfort was notably strong -- participants responded positively to knowing all inference ran locally, suggesting on-device deployment contributes to trust beyond its technical merits. Confidence and trust in the system's proactive interventions were broadly positive, with some residual uncertainty expected from first-time users encountering a novel interaction pattern. Critically, participants did not find the system mentally taxing, a direct validation of the agent's deliberate restraint and the Progressive Disclosure strategy, which together kept cognitive load low throughout the task. Response latency was acceptable to most users, with the immediate verbal acknowledgment and waiting sound effectively masking the 8–12 second VLM verification delay for the majority of participants, though it remains a target for future optimization.

Future work should focus on broader task domains, multi-camera setups to address occlusion, backward revision of completed steps, and reduced VLM latency. \sys ultimately shows that useful AI assistance in the physical world does not require the cloud -- it requires the right design.

\bibliographystyle{ACM-Reference-Format}
\bibliography{references}

\appendix
\clearpage
\section{Multi-Judge LLM Evaluation Prompt}
\label{appendix:llm_as_judge}
\begin{table}[h]
\caption{}
\resizebox{\linewidth}{!}{
\begin{tabular}{@{}l@{}}
\toprule[2pt]
\begin{tabular}[c]{@{}l@{}}You are an expert evaluator for an AI-assisted furniture assembly system. Your task is to evaluate the quality of an AI assistant's response given:\\ 1. A knowledge base (KB) describing the assembly process\\ 2. The current assembly state\\ 3. A user question\\ 4. The assistant`s answer\\ \textbf{INPUTS}\\ \qquad Knowledge Base (KB): \{knowledge\_base\}\\ \qquad Ground Truth Sequence: \{ground\_truth\_sequence\}\\ \qquad Task Sequence (may be incomplete or user-facing version): \{task\_sequence\}\\ \qquad Rules for Completion: \{rules\_for\_completion\}\\ \qquad Current Assembly State: \{current\_state\}\\ \qquad User Question: \{question\}\\ \qquad Assistant Answer: \{answer\}\\  \textbf{EVALUATION CRITERIA}\\ Evaluate the assistant`s answer along the following three axes:\\ \textbf{1. Hallucination (0–5)}\\ Does the answer introduce information that is NOT supported by the KB or current state?\\ \qquad 5 → No hallucination (fully grounded in KB/state)\\ \qquad 4 → Minor unsupported assumptions, but mostly grounded\\ \qquad 3 → Some unsupported or speculative content\\ \qquad 2 → Significant hallucination affecting reliability\\ \qquad 1 → Mostly hallucinated\\ \qquad 0 → Completely fabricated / unrelated\\ \textbf{2. Correctness (0–5)}\\ Is the answer factually correct and consistent with the KB and current state?\\ \qquad 5 → Fully correct and consistent\\ \qquad 4 → Minor inaccuracies but overall correct\\ \qquad 3 → Partially correct with notable mistakes\\ \qquad 2 → Mostly incorrect\\  \qquad 1 → Largely incorrect\\ \qquad 0 → Completely wrong\\  \textbf{3. Helpfulness (0–5)}\\ Does the answer effectively help the user progress in assembly?\\ \qquad 5 → Clear, actionable, and directly helps progress\\ \qquad 4 → Helpful but slightly incomplete or unclear\\ \qquad 3 → Somewhat helpful but lacks clarity or steps\\  \qquad 2 → Minimally helpful\\  \qquad 1 → Confusing or not actionable\\  \qquad 0 → Not helpful at all\\ \textbf{IMPORTANT INSTRUCTIONS}\\ * Ground your evaluation strictly in the provided KB and current state.\\ * Do NOT assume missing steps unless clearly implied.\\ * Penalize hallucinations even if the answer sounds plausible.\\ * Consider whether the answer is appropriate for the ``current assembly stage''.\\ \textbf{OUTPUT FORMAT (STRICT JSON)}\\ \{\{\\ \qquad "hallucination\_score": \textless{}0-5\textgreater{},\\ \qquad  "correctness\_score": \textless{}0-5\textgreater{},\\  \qquad  "helpfulness\_score": \textless{}0-5\textgreater\\ \}\}\end{tabular} \\ \bottomrule[2pt]
\end{tabular}
}
\end{table}
\section{Hyper-parameter Search}
\label{appendix:hyperparameter_search}

\begin{table*}[h]
\centering
\caption{}
\label{tab:hyper_parameter_ablation}
\definecolor{best}{gray}{0.9} 
\resizebox{\linewidth}{!}{
\begin{tabular}{ccc|cc|cc|cc}
\hline
 & & & \multicolumn{2}{c|}{Single Frame} & \multicolumn{2}{c|}{Multi Frame (0-100\%)} & \multicolumn{2}{c}{Multi Frame (50-100\% RWS)} \\
W & S & Check & Acc (\%) & Adh (\%) & Acc (\%) & Adh (\%) & Acc (\%) & Adh (\%) \\ \hline
60 & 20 & None & 77.00 $\pm$ 12.74 & 35.24 $\pm$ 29.47 & 75.62 $\pm$ 11.35 & 40.00 $\pm$ 43.33 & 76.25 $\pm$ 11.82 & 40.00 $\pm$ 43.33 \\
60 & 20 & 15 secs & 72.14 $\pm$ 8.78 & 52.38 $\pm$ 32.12 & 76.16 $\pm$ 11.34 & 43.81 $\pm$ 33.06 & 74.52 $\pm$ 11.23 & 43.81 $\pm$ 33.06 \\
60 & 20 & 20 secs & 72.65 $\pm$ 10.09 & 49.52 $\pm$ 27.06 & 73.47 $\pm$ 9.38 & 58.10 $\pm$ 34.07 & 74.52 $\pm$ 11.23 & 58.10 $\pm$ 34.07 \\
60 & 20 & 25 secs & 74.21 $\pm$ 12.07 & 52.38 $\pm$ 24.97 & 74.68 $\pm$ 12.16 & 60.95 $\pm$ 21.93 & 74.01 $\pm$ 11.95 & 60.95 $\pm$ 21.93 \\ \hline
30 & 15 & None & 75.64 $\pm$ 7.85 & 68.57 $\pm$ 35.57 & 64.54 $\pm$ 37.47 & 45.71 $\pm$ 39.64 & 61.88 $\pm$ 35.81 & 48.57 $\pm$ 39.89 \\
30 & 15 & 15 secs & 77.55 $\pm$ 9.74 & 56.19 $\pm$ 29.82 & 80.37 $\pm$ 11.90 & 44.76 $\pm$ 30.04 & 75.50 $\pm$ 8.87 & 62.86 $\pm$ 29.62 \\
\textbf{30} & \textbf{15} & \textbf{20 secs} & 78.25 $\pm$ 9.49 & 64.76 $\pm$ 25.78 & 80.37 $\pm$ 11.90 & 56.19 $\pm$ 21.92 & 75.82 $\pm$ 8.89 & \cellcolor{best}\textbf{74.28 $\pm$ 27.48} \\
30 & 15 & 25 secs & 78.29 $\pm$ 9.88 & 64.76 $\pm$ 23.71 & 80.37 $\pm$ 11.90 & 59.05 $\pm$ 18.00 & 75.73 $\pm$ 9.67 & 68.57 $\pm$ 25.55 \\ \hline
20 & 10 & None & 77.87 $\pm$ 10.34 & 60.00 $\pm$ 30.97 & 75.80 $\pm$ 10.96 & 31.43 $\pm$ 34.10 & 75.40 $\pm$ 10.20 & 49.05 $\pm$ 35.30 \\
20 & 10 & 15 secs & 77.93 $\pm$ 9.89 & 51.43 $\pm$ 34.40 & 77.27 $\pm$ 8.76 & 43.66 $\pm$ 39.16 & 77.22 $\pm$ 8.11 & 48.57 $\pm$ 37.25 \\
20 & 10 & 20 secs & 76.75 $\pm$ 9.46 & 60.00 $\pm$ 38.33 & 75.52 $\pm$ 8.24 & 66.51 $\pm$ 25.12 & 77.22 $\pm$ 8.11 & 60.00 $\pm$ 36.98 \\
20 & 10 & 25 secs & 78.64 $\pm$ 10.47 & 60.95 $\pm$ 26.17 & 75.92 $\pm$ 8.27 & 66.51 $\pm$ 25.12 & 77.22 $\pm$ 8.11 & 71.43 $\pm$ 28.57 \\ \hline
10 & 5 & None & 79.24 $\pm$ 9.76 & 62.86 $\pm$ 37.25 & 78.11 $\pm$ 10.15 & 49.52 $\pm$ 32.23 & 80.76 $\pm$ 11.14 & 49.52 $\pm$ 32.23 \\
10 & 5 & 15 secs & 79.70 $\pm$ 9.92 & 54.29 $\pm$ 32.57 & 78.81 $\pm$ 10.19 & 49.37 $\pm$ 37.14 & 78.27 $\pm$ 9.87 & 54.29 $\pm$ 32.57 \\
10 & 5 & 20 secs & 79.95 $\pm$ 9.56 & 62.86 $\pm$ 37.25 & 78.88 $\pm$ 10.43 & 60.80 $\pm$ 37.92 & 78.18 $\pm$ 9.68 & 62.86 $\pm$ 34.40 \\
10 & 5 & 25 secs & 79.71 $\pm$ 9.94 & 74.29 $\pm$ 25.55 & 78.85 $\pm$ 10.45 & 72.23 $\pm$ 27.61 & 78.33 $\pm$ 9.63 & 74.29 $\pm$ 29.28 \\ \hline
5 & 2 & None & 79.56 $\pm$ 10.97 & 43.34 $\pm$ 34.52 & 80.52 $\pm$ 11.05 & 44.76 $\pm$ 37.58 & 83.98 $\pm$ 13.11 & 40.95 $\pm$ 35.25 \\
5 & 2 & 15 secs & 79.80 $\pm$ 10.86 & 31.91 $\pm$ 18.10 & 80.42 $\pm$ 10.63 & 38.10 $\pm$ 21.30 & 80.30 $\pm$ 11.12 & 38.10 $\pm$ 21.30 \\
5 & 2 & 20 secs & 80.35 $\pm$ 10.67 & 43.81 $\pm$ 33.06 & 80.49 $\pm$ 10.68 & 43.81 $\pm$ 33.06 & 80.33 $\pm$ 10.85 & 43.81 $\pm$ 33.06 \\
5 & 2 & 25 secs & 79.81 $\pm$ 10.74 & 40.48 $\pm$ 35.15 & 80.43 $\pm$ 10.60 & 56.19 $\pm$ 31.48 & 80.14 $\pm$ 10.96 & 47.47 $\pm$ 31.94 \\ \hline
2 & 1 & None & 79.88 $\pm$ 10.59 & 22.86 $\pm$ 21.66 & 84.75 $\pm$ 13.29 & 35.24 $\pm$ 16.01 & 79.69 $\pm$ 10.98 & 25.71 $\pm$ 21.19 \\
2 & 1 & 15 secs & 79.67 $\pm$ 10.24 & 25.71 $\pm$ 21.19 & 80.14 $\pm$ 11.11 & 22.86 $\pm$ 16.29 & 79.71 $\pm$ 10.25 & 25.71 $\pm$ 21.19 \\
2 & 1 & 20 secs & 79.71 $\pm$ 10.34 & 25.71 $\pm$ 21.19 & 79.82 $\pm$ 10.88 & 22.86 $\pm$ 16.29 & 79.45 $\pm$ 10.09 & 25.71 $\pm$ 21.19 \\
2 & 1 & 25 secs & 79.39 $\pm$ 9.98 & 25.71 $\pm$ 21.19 & 80.08 $\pm$ 11.07 & 22.86 $\pm$ 16.29 & 79.78 $\pm$ 10.51 & 25.71 $\pm$ 21.19 \\ \hline
1 & 1 & None & 80.36 $\pm$ 10.91 & 17.15 $\pm$ 15.65 & 80.49 $\pm$ 11.22 & 25.72 $\pm$ 18.63 & 80.42 $\pm$ 11.01 & 25.72 $\pm$ 18.63 \\
1 & 1 & 15 secs & 80.28 $\pm$ 11.23 & 22.86 $\pm$ 16.29 & 79.90 $\pm$ 10.88 & 20.00 $\pm$ 16.29 & 80.02 $\pm$ 10.98 & 22.86 $\pm$ 16.29 \\
1 & 1 & 20 secs & 80.35 $\pm$ 11.24 & 22.86 $\pm$ 16.29 & 80.13 $\pm$ 11.16 & 22.86 $\pm$ 16.29 & 80.10 $\pm$ 11.02 & 20.00 $\pm$ 16.29 \\
1 & 1 & 25 secs & 80.17 $\pm$ 11.11 & 25.72 $\pm$ 18.63 & 80.10 $\pm$ 11.10 & 25.72 $\pm$ 18.63 & 80.31 $\pm$ 11.26 & 25.72 $\pm$ 18.63 \\
\hline
\end{tabular}
}
\end{table*}

\section{Atomic Steps Extraction Prompt}
\label{appendix:step_extract_prompt}

\begin{table}[H]

\caption{}
\resizebox{\linewidth}{!}{
\begin{tabular}{@{}l@{}}
\toprule[2pt]
\begin{tabular}[c]{@{}l@{}}\textbf{ROLE}\\ Act as an expert Furniture Assembly Action Classifier.\\ \textbf{VIDEO SPECIFICATIONS}\\ \qquad Duration:  180 seconds\\ \qquad Frame Rate: 30 FPS\\ \qquad \textbf{Processing Logic:} Every 1 second of video equals 30 frames. Use this linear scale to ensure timestamps and frame indices do not drift over the 3-minute duration.\\ \textbf{TASK}\\ Analyze the provided video and extract a sequence of atomic steps.\\  \textbf{CONSTRAINTS}\\ \qquad 1. ONLY use the following standardized action names (Verb-Noun pairs):\\   \qquad \qquad - Verbs: {[}pick up, align, spin, attach, flip, insert, tighten, remove, align and spin{]}\\  \qquad \qquad  - Nouns: {[}leg, shelf, table, screw, base, panel, drawer, frame{]}\\ \qquad 2. Cardinality Rule: Log EVERY instance of a repetitive action. Do NOT summarize.\\ \qquad 3. If an action does not fit the list, use the label "others".\\ \qquad 4. Follow the Step Extraction Protocol: split complex movements into atomic steps.\\ \textbf{OUTPUT FORMAT (JSON)}\\ Return ONLY the JSON.\\ \{\\ \qquad   "reference\_video\_id": "\{\{job\_id\}\}",\\  \qquad  "furniture\_name": "...",\\  \qquad "metadata": \{\\  \qquad   "fps": 30,\\  \qquad   "video\_duration\_seconds": 180\\   \},\\   "steps": {[}\\     \{\\     \qquad  "name": "verb + noun",\\    \qquad   "timestamp": "MM:SS",\\   \qquad    "frame\_index": 0000,\\    \qquad   "description": "Short visual anchor (e.g., hand touches leg)"\\     \}\\   {]}\\ \}\end{tabular} \\ \bottomrule[2pt]
\end{tabular}
}
\end{table}
\section{Video Dataset for Ablation Studies}
\label{appendix:video_data_list}

\begin{table}[h]
\centering
\caption{List of 15 Video Sequences used in Ablation Studies}
\label{tab:video_dataset}
\begin{tabular}{ll}
\hline
\textbf{Furniture Category} & \textbf{Video Sequence Identifier} \\ \hline
\textit{Lack Coffee Table} & \texttt{0001\_black\_floor\_01\_01\_2019\_08\_14\_15\_22} \\
& \texttt{0001\_black\_floor\_05\_02\_2019\_08\_19\_16\_51} \\
& \texttt{0005\_white\_table\_10\_04\_2019\_08\_28\_14\_45} \\
& \texttt{0014\_black\_table\_02\_01\_2019\_08\_16\_13\_14} \\
& \texttt{0016\_oak\_table\_07\_03\_2019\_08\_21\_17\_08} \\ \hline
\textit{Lack Side Table} & \texttt{0001\_white\_table\_02\_01\_2019\_08\_16\_13\_54} \\
& \texttt{0002\_oak\_floor\_01\_01\_2019\_08\_12\_16\_33} \\
& \texttt{0003\_white\_floor\_01\_01\_2019\_08\_14\_14\_49} \\
& \texttt{0004\_white\_floor\_01\_01\_2019\_08\_14\_16\_11} \\
& \texttt{0005\_oak\_table\_07\_03\_2019\_08\_21\_15\_55} \\ \hline
\textit{Lack TV Bench} & \texttt{0001\_black\_floor\_01\_01\_2019\_08\_13\_13\_19} \\
& \texttt{0001\_white\_table\_02\_01\_2019\_08\_16\_13\_57} \\
& \texttt{0003\_white\_floor\_01\_01\_2019\_08\_14\_14\_58} \\
& \texttt{0008\_white\_floor\_01\_01\_2019\_08\_15\_11\_06} \\
& \texttt{0025\_black\_table\_04\_02\_2019\_08\_20\_13\_48} \\ \hline
\end{tabular}
\end{table}

\end{document}